\documentclass[10pt,twocolumn,letterpaper]{article}

\usepackage{cvpr}              % To produce the CAMERA-READY version
\usepackage[dvipsnames]{xcolor}

\definecolor{cvprblue}{rgb}{0.21,0.49,0.74}
\usepackage[pagebackref,breaklinks,colorlinks,citecolor=cvprblue]{hyperref}

\usepackage{multirow}
\usepackage{makecell}
\usepackage{hhline}
\usepackage{bbding}
\usepackage{booktabs}

\usepackage{arydshln}
\usepackage{amssymb}

\def\paperID{15163} % *** Enter the Paper ID here
\def\confName{CVPR}
\def\confYear{2024}

\usepackage{color}

\title{PLACE: Adaptive Layout-Semantic Fusion for Semantic Image Synthesis}

\author{Zhengyao Lv$^1$, Yuxiang Wei$^2$, Wangmeng Zuo$^{2,3}$, Kwan-Yee K. Wong$^{1(}$\Envelope$^)$\\
		$^1$The University of Hong Kong
        $^2$Harbin Institute of Technology
        $^3$Pazhou Lab, Guangzhou
        \\
		{\tt\small \{cszy98, yuxiang.wei.cs\}@gmail.com wmzuo@hit.edu.cn kykwong@cs.hku.hk}
	}

\begin{document}
\maketitle
\begin{abstract}
Recent advancements in large-scale pre-trained text-to-image models have led to remarkable progress in semantic image synthesis.
Nevertheless, synthesizing high-quality images with consistent semantics and layout remains a challenge.
In this paper, we propose the adaPtive LAyout-semantiC fusion modulE (PLACE) that harnesses pre-trained models to alleviate the aforementioned issues. 
Specifically, we first employ the layout control map to faithfully represent layouts in the feature space. 
Subsequently, we combine the layout and semantic features in a timestep-adaptive manner to synthesize images with realistic details.
During fine-tuning, we propose the Semantic Alignment (SA) loss to further enhance layout alignment. Additionally, we introduce the Layout-Free Prior Preservation (LFP) loss, which leverages unlabeled data to maintain the priors of pre-trained models, thereby improving the visual quality and semantic consistency of synthesized images.
Extensive experiments demonstrate that our approach performs favorably in terms of visual quality, semantic consistency, and layout alignment.
The source code and model are available at \href{https://github.com/cszy98/PLACE/tree/main}{PLACE}. 
\end{abstract}
\section{Introduction}
\label{sec:introv1}
Semantic image synthesis aims to generate high-quality images that are aligned with given semantic maps. 
It provides users the flexibility to precisely control the spatial layout of synthesized images using semantic maps while having important applications in content creation~\cite{chen2017photographic,zhu2020sean,goel2023pair}, image editing~\cite{ntavelis2020sesame,luo2022context,luo2023siedob}, and data augmentation~\cite{yang2023freemask}.
\begin{figure}[t]
\centering
\vspace{-10pt}
\includegraphics[width=.95\linewidth]{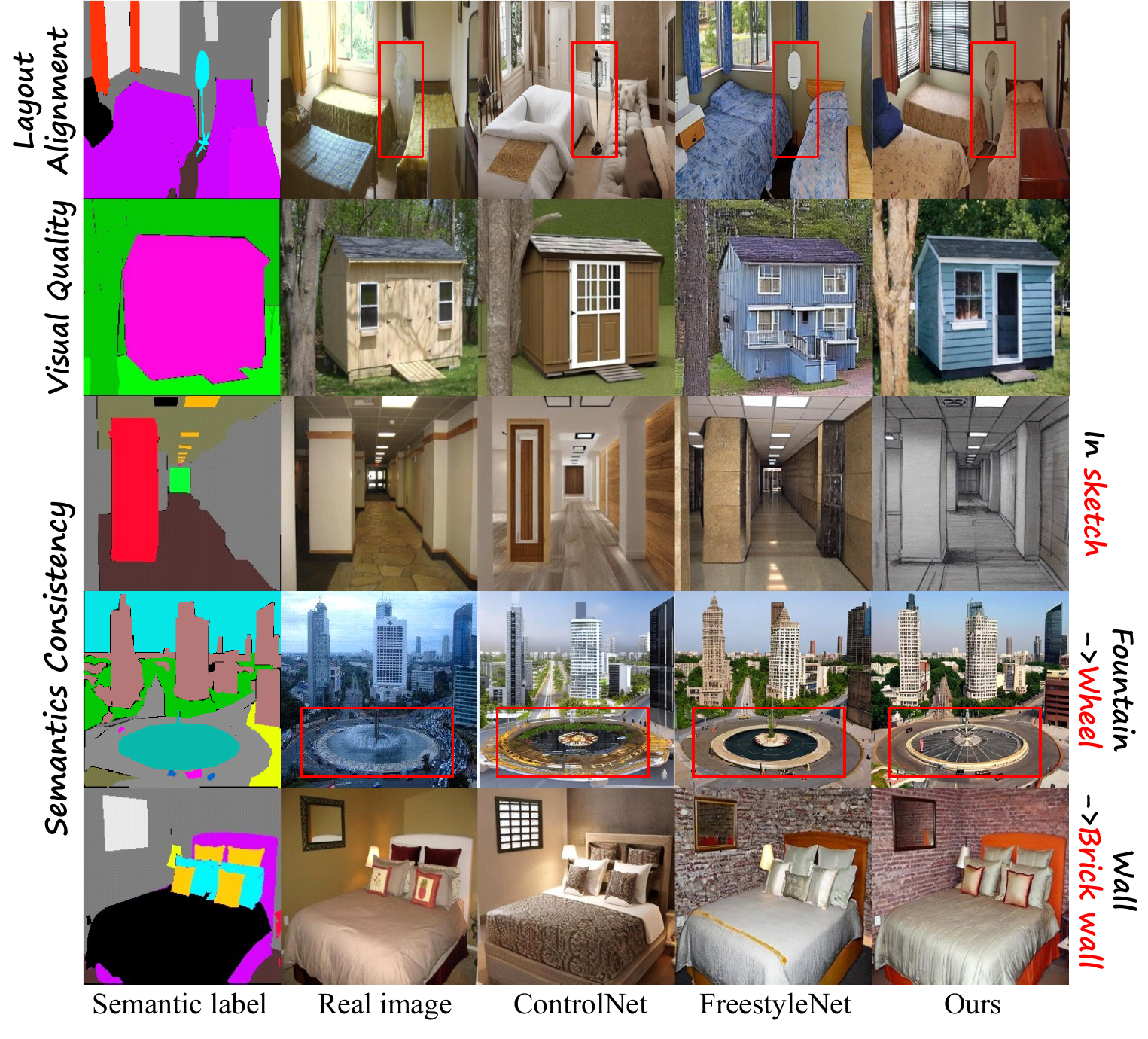}
\vspace{-13pt}
\caption{Comparisons in terms of visual quality as well as layout alignment and semantic consistency. Zoom in for details.}
\label{fig:compare}
\vspace{-20pt}
\end{figure}

Earlier semantic image synthesis works~\cite{isola2017image,park2019semantic,liu2019learning,sushko2020you} mainly relied on Generative Adversarial Networks (GANs)~\cite{goodfellow2014generative} and trained a model using semantic maps as condition within specific domains.
However, due to the limited scale of the training dataset, the quality and diversity of generated images are usually compromised. 
Recently, large-scale text-to-image models~\cite{ramesh2022hierarchical,rombach2022high,saharia2022photorealistic} have shown high-quality and diverse generation results with open-vocabulary textual prompts.
Based on these pre-trained text-to-image models (\eg, Stable Diffusion~\cite{rombach2022high}), ControlNet~\cite{zhang2023adding} and T2I-Adapter~\cite{mou2023t2i} introduced an additional adapter to inject layout guidance for high-quality semantic image synthesis.
Nevertheless, these adapters failed to integrate textual semantics with corresponding regions accurately, resulting in inconsistent layouts in generated results, as shown in Fig.~\ref{fig:compare}.

%
% rca --attend
To facilitate the layout consistency, FreestyleNet~\cite{xue2023freestyle} proposed an RCA module that forces each intermediate image token to attend to the respective textual semantic, while fine-tuning the diffusion model with RCA. 
However, the semantic map used in RCA is directly adapted to the intermediate image features in latent diffusion, which is considerably smaller than its original size (\eg, $64 \times 64$ compared to $512\times 512$), leading to inevitable layout information loss.
Moreover, the mechanism of RCA disrupts the global interaction between image and text tokens, impeding the synthesis of high-quality images.

\begin{figure*}[!t]
\vspace{-15pt}
\centering
\includegraphics[width=.88\linewidth]{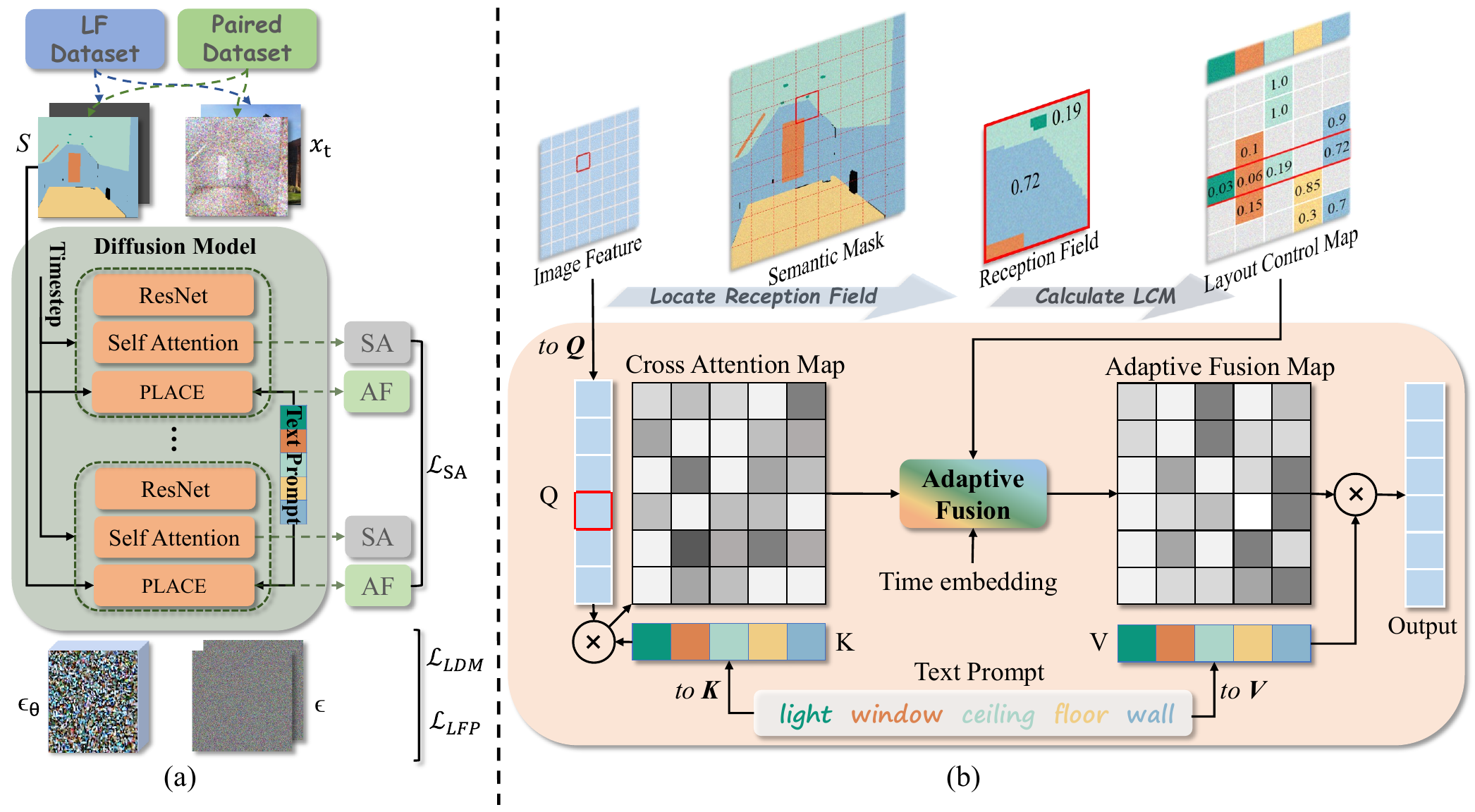}
\vspace{-14pt}	
\caption{Overview of our method. (a) We utilize the layout control map calculated from semantic map $S$ and PLACE for layout control. During fine-tuning, we combine the $\mathcal{L}_{LDM}$, $\mathcal{L}_{SA}$, and $\mathcal{L}_{LFP}$ as optimization objective. (b) Calculation of the layout control map and details of the adaptive layout-semantic fusion module. Each vector in the Layout Control Map encodes all the semantic components in the reception field. The adaptive layout-semantic fusion module blends the layout and semantics feature in a timestep-adaptive way.}
\label{fig:overview}
\vspace{-14pt}	
\end{figure*}

To alleviate the aforementioned issues, we propose the ada{P}tive {LA}yout-semanti{C} fusion modul{E} (termed as {PLACE}) as depicted in Fig.~\ref{fig:overview}, which leverages pre-trained Stable Diffusion for high-quality and faithful semantic image synthesis. 
Firstly, inspired by the spatio-textual representation~\cite{avrahami2023spatext}, we introduce the layout control map (LCM), which represents the layout information faithfully in low-resolution feature space. 
Specifically, we explore the proportion of each semantic component within the receptive field of each image token in the intermediate image features and utilize a vector composed of these proportions as the layout feature for this image token. 
Such a layout control map retains layout information accurately in the feature space and can be then incorporated with textual features to guide the semantic image synthesis. 

Manually constraining the regions influenced by each semantic component with the layout control map allows for the manipulation of the layout of synthesized images.
However, we noticed that it also restricts the interaction between image tokens and global text tokens, compromising the visual quality of synthesized details.
To effectively integrate layout control maps, while preserving the beneficial interactions for better visual quality, we develop a timestep-adaptive layout-semantic fusion module. Specifically, for each fusion module, a time-adaptive fusion parameter is learned from time embedding. Subsequently, this parameter is employed to adaptively combine our layout control map with the original cross-attention maps which encapsulate the global semantics.
The resulting adaptive fusion maps not only encompass faithful layout information but also maintain the influence of contextual textual tokens, thereby improving the visual quality of generated images.

Additionally, we propose effective {S}emantic {A}lignment \linebreak ({SA}) loss and {L}ayout-{F}ree {P}rior {P}reservation loss to facilitate the fine-tuning.
The SA loss constrains the weighted aggregation results of the adaptive fusion map and self-attention maps to be as close as possible to the original adaptive fusion maps. It enhances the internal interactions of image tokens within the same or related semantic region, consequently improving the layout consistency and visual quality.
Due to the limited scale of datasets, priors of the pre-trained model are prone to perturbation while fine-tuning. 
Our proposed LFP loss helps preserve priors without involving layout annotation during fine-tuning.
Specifically, we compute the denoising loss in a layout-free manner with the text-image pairs to preserve the semantic concepts embedded in the pre-trained model.
Owing to the enhanced preservation and utilization of semantic priors, our method exhibits better visual quality and semantic consistency, even in new domains (as shown in Fig.~\ref{fig:compare}).

The contributions of this work can be summarized as:
	\begin{itemize}
	\setlength{\itemsep}{0pt}
	\setlength{\parsep}{0pt}
	\item We introduce the layout control map as a reliable layout representation and propose an adaptive layout-semantic fusion module to adaptively integrate the layout and semantic features for semantic image synthesis.
        \item We propose effective SA and LFP losses. The former enhances the layout consistency of generated images, while the latter helps preserve the semantic priors of pre-trained models with readily available text-image pairs.
	\end{itemize}

\section{Related Work}

\vspace{-2pt}
\subsection{Semantic Image Synthesis}
Semantic image synthesis aims to synthesize realistic images with given semantic masks. Previous works primarily achieved the layout control over generated images through Generative Adversarial Networks (GANs)~\cite{goodfellow2014generative}. 

Pix2pix~\cite{isola2017image} was the first to propose using an encoder-decoder generator and a PatchGAN discriminator for semantic image synthesis. Pix2pixHD~\cite{wang2018high} accomplished high-resolution image synthesis by employing a coarse-to-fine generator and multi-scale discriminators. SPADE~\cite{park2019semantic} proposed using spatially adaptive transformations learned from semantic maps to modulate features and significantly improved the image quality. Subsequently, CC-FPSE~\cite{liu2019learning} introduced predicting conditional convolution kernel parameters based on semantic layouts and utilized a feature pyramid semantic-embedding discriminator to encourage the generator to produce images with higher-quality details and better semantic alignment. More recently, SCGAN~\cite{wang2021image} learned a semantic vector to parameterize conditional convolution kernels and normalization parameters. LGGAN~\cite{tang2020local} introduced the utilization of a local class-specific and global image-level generative adversarial network to individually learn the appearance distribution of each object category and the global image. OASIS~\cite{sushko2020you} innovatively designed a segmentation network-based discriminator, providing the generator with more potent feedback, thereby generating semantically aligned images with higher fidelity. Besides, there are also some methods~\cite{qi2018semi,shi2022retrieval,lv2022semantic, wei2023inferring, tang2023edge} exploring structural and shape information in semantic maps to enhance the quality of images.

Despite significant achievements of previous methods in semantic image synthesis, generated images still show limited quality and diversity due to constraints in the scale of training data and the representation of semantic layouts.

\vspace{-4pt}
\subsection{Layout Controllable Text-to-Image Synthesis}

Text-to-image synthesis focuses on generating images conditioned on given text prompts. Benefiting from the powerful diffusion model~\cite{ho2020denoising, song2020score} and extensive text-image training data, text-to-image synthesis has achieved unprecedented successes in terms of image quality, diversity, and alignment with the provided text~\cite{nichol2021glide,ramesh2022hierarchical,rombach2022high,saharia2022photorealistic}.  
Among them, the balance between efficiency and quality of LDM~\cite{rombach2022high} has attracted significant attention, making it the foundational model for many controllable~\cite{zhang2023adding, zhao2024uni} or customized image synthesis works~\cite{ruiz2023dreambooth,wei2023elite, hao2023vico}.

Subsequent works investigated the utilization of pre-trained models to achieve layout-controllable text-to-image synthesis~\cite{avrahami2023spatext,xiao2023r,kim2023dense,xie2023boxdiff,phung2023grounded}. The eDiff-I~\cite{balaji2022ediffi} and the Two Layout Guidance~\cite{chen2023training} iteratively optimize the alignment between the constrained cross-attention map and the target layout. Nevertheless, they can merely roughly control the positioning of the synthesized objects. 
ControlNet~\cite{zhang2023adding} and T2I-Adapter~\cite{mou2023t2i} encode semantic maps with an additional layout encoder. However, constrained by the generalization capacity of the layout encoder, they fail to overcome the limitations of layout consistency. Another category of methods controls the layout of synthesized images in a training-free manner. FreestyleNet~\cite{xue2023freestyle} introduces Rectified Cross Attention (RCA) to replace the cross attention module in Stable Diffusion, enabling each text token to interact exclusively with the corresponding image feature region. Subsequently, the pre-trained Stable Diffusion model is fine-tuned on specific domains to adapt to RCA. FreestyleNet has made progress in semantic consistency and layout alignment. However, due to the loss of layout information when utilizing semantic maps with RCA, the generated images lack sufficient layout alignment. Furthermore, due to the modifications in cross-attention and the limited scale of fine-tuning datasets, FreestyleNet is prone to losing priors in the pre-trained model and still exhibits limitations in visual quality and semantic consistency.

\vspace{-1em}
\section{Proposed Method}
\vspace{-0.5em}

Given a semantic map $S \in \mathbb{R}^{H\times W\times C}$ with $C$ semantic classes semantic image synthesis aims to synthesize photo-realistic images that are well aligned with $S$. The value of $C$ is determined by the number of semantic categories specified by the user, rather than the cardinality of a predefined closed set.
To enable controllable image synthesis with desired layouts, we first employ a faithful layout control map as layout representations in feature space. 
Then, we propose the PLACE to adaptively integrate the layout and semantic features, as illustrated in Fig.~\ref{fig:overview}~(b). 
During fine-tuning, we further introduce the semantic alignment~(SA) loss to enhance the layout alignment and the layout-free prior preservation~(LFP) loss to improve the performance of visual quality and semantic consistency. 
In the following subsections, we first give a concise introduction of the pre-trained text-to-image model we employed, namely Stable Diffusion~\cite{rombach2022high}. 
We then provide the details of our PLACE and learning objective.

\vspace{-4pt}
\subsection{Preliminary: Stable Diffusion}

Stable Diffusion~\cite{rombach2022high} is a text-to-image synthesis model based on the diffusion process in the latent space. 
It comprises two components, namely an autoencoder and a conditional latent diffusion model (LDM). 
The autoencoder $\mathcal{E}$ is designed to learn a latent space that is perceptually equivalent to the image space.
Meanwhile, the conditional LDM $\epsilon_{\theta}$ is parameterized as a UNet with cross-attention and trained on a large-scale dataset of text-image pairs via:
\begin{equation}
\mathcal{L}_{LDM}:=\mathbb{E}_{\mathcal{E}(x),y,\epsilon \sim\mathcal{N}(0,1),t}[||\epsilon - \epsilon_{\theta}(z_t,t,\tau_{\theta}(y))||^2_2],
\label{eq:sd_ldm}
\end{equation}
where $\epsilon$ is the target noise, $\tau_{\theta}$ and $y$ are the pretrained CLIP~\cite{radford2021learning} text encoder and text prompts, respectively, and $z_t$ is the noisy latent at timestep $t$. 

In the conditional LDM, the text feature is integrated into the intermediate layers of the UNet through the cross-attention modules:
\begin{equation}
\begin{aligned}
\ \ Q = W_Q \cdot\phi_i,\ K = W_K \cdot \tau_{\theta}(y),\ V = W_V \cdot \tau_{\theta}(y), \\ \text{Attention}(Q,K,V)=\text{softmax}(\frac{QK^T}{\sqrt{d}})V,
\end{aligned}
\end{equation}
where $\phi_i$ is the intermediate image features.
$W_Q$, $W_K$, and $W_V$ denote the learnable projection matrices of query, key, and value, respectively. 
In the cross-attention module, the interaction between each text token and image token significantly influences the layout of the generated image. 
%Next, we will demonstrate how our XXNet integrates layout with semantic information to achieve controllable layout image synthesis.

\begin{figure}[t]
\vspace{-1pt}
\centering
\includegraphics[width=0.94\linewidth]{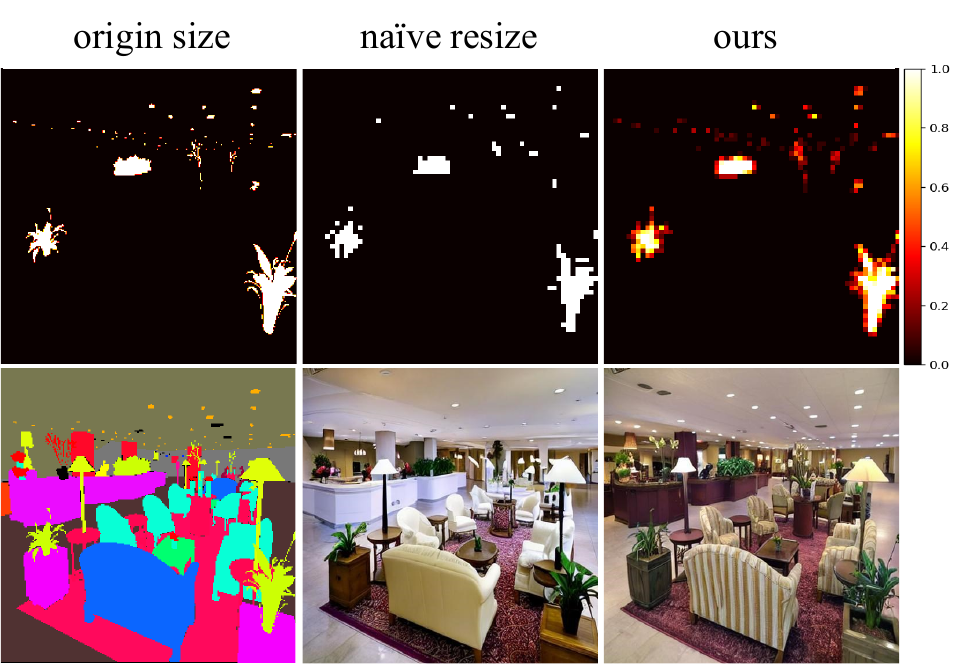}
\vspace{-6pt}
\caption{Comparison between naive resize and layout control map regarding information preservation~(downsampling by 8 times). The $1st$ column displays the original mask of `plants, lights' and full semantic map, the $2nd$ column shows the nearest resized mask and corresponding synthesized image, and the $3rd$ column presents the representation of the layout control map and its generated image. A higher value indicates a higher proportion of semantics within its patch. Ours preserves more details.}
\label{fig:fig3}
\vspace{-16pt}
\end{figure}

\vspace{-6pt}
\subsection{Adaptive Layout-Semantic Fusion Network}

In this subsection, we will first introduce our proposed layout control map and then present the details of our adaptive layout-semantic fusion module.

\noindent \textbf{Layout control map.} 
It has been noted that cross-attention maps in Stable Diffusion are closely related to the layout of the synthesized image~\cite{balaji2022ediffi,avrahami2023spatext}. 
Specifically, $A^{ca}_{i,j}$ in the cross-attention map $A^{ca}\in \mathbb{R}^{(HW)\times N}$ determines the strength of the association between the $i$-th image token and the $j$-th text token, thus influencing the layout of synthesized images.
Previous works roughly controlled the position of specific objects in the synthesized image by constraining the influence region of specific tokens in the cross-attention map.
However, due to the significantly smaller size of the intermediate image features in LDM (less than or equal to $64\times64$) compared to that of given semantic layouts ($512\times512$ or larger), simply resizing semantic maps to adapt to the size of intermediate features inevitably leads to distortion or even loss of details. 
For example, as illustrated in Fig.~\ref{fig:fig3}, even when the semantic map is resized to only $1/8$ of its original size with a naive nearest-neighbor interpolation, the details of the `plants' are distorted, and some instances of `light' are lost. 
Moreover, the image features from deeper layers have smaller dimensions, making it challenging to synthesize images that align precisely with the given semantic maps.

To address the above issue, we propose a layout control map that encodes layout information in the low-resolution feature space with less loss of layout information. 
For each token of the intermediate image features, we investigate all the semantic components within its receptive field, along with the proportion occupied by each class. 
We then use a vector composed of these proportions as the layout feature for this token. 
As shown in the top of Fig.~\ref{fig:overview} (b), within the receptive field of the image token selected by the red border, there are four semantic categories, namely `wall', `ceiling', `window', and `light', and each corresponding to a different proportion. 
The vector formed by the proportions faithfully encodes the layout information within the receptive field of this image token. 
Given a semantic map $\hat{S}\in \mathbb{R}^{(HW)\times C}$ reshaped from $S\in \mathbb{R}^{H\times W\times C}$, the calculation of layout control map $L\in \mathbb{R}^{(hw)\times N}$ can be formulated as following:
\begin{equation}
L_{i,j}=
\begin{cases}
%\frac{|\hat{S}_{RF(i),S(j)}=1|}{HW/hw}& \text{$|\hat{S}_{RF(i),S(j)}=1|\neq 0$}\\
\frac{|\hat{S}_{RF(i),S(j)}=1|}{|{RF(i)}|},& \text{$|\hat{S}_{RF(i),S(j)}=1|\neq 0$}\\
-\infty, & \text{otherwise},
\end{cases}
\end{equation}
where $RF(i)$ denotes the receptive field of the $i$-th image token and $S(j)$ is the corresponding semantic channel of the $j$-th text token. 
$|\cdot|$ represents the number of elements in the set. 
As can be observed from Fig.~\ref{fig:fig3}, our layout control map encodes faithful layout details, including the branches and leaves of the plants, and subtle lighting, resulting in a synthesized image with richer and accurate details.
\begin{figure*}[t]
\vspace{-1pt}
\centering
\includegraphics[width=1.0\linewidth]{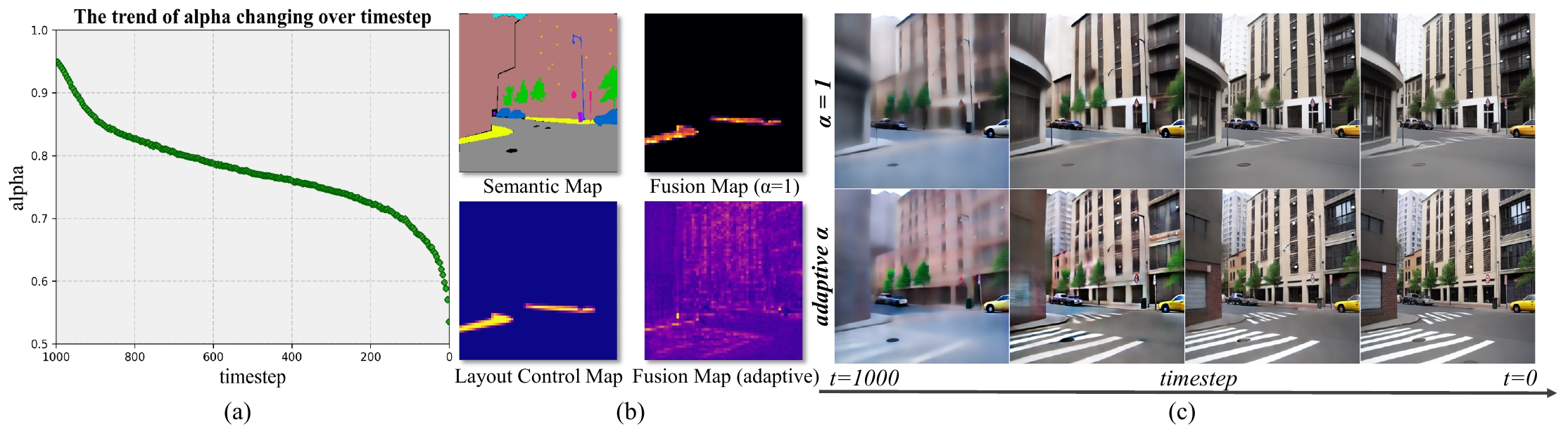}
\vspace{-22pt}
\caption{Analysis of adaptive fusion module. (a) shows the variation of adaptive $\alpha$ with respect to the timestep. The $\alpha$ decreases as the timestep progresses. (b) presents the layout control map of the `sidewalk' and the corresponding comparison of the fusion maps~(at $t=800/1000$) between fixed $\alpha=1$ and adaptive $\alpha$. (c) illustrates the variation of predicted $\hat{x}_0$ with respect to the sampling steps: one with a fixed $\alpha$ and the other with an adaptive $\alpha$. The latter leads to the synthesis of more realistic details. Zoom in for details.}
\label{fig:fig4}
\vspace{-12pt}
\end{figure*}

\noindent \textbf{Adaptive fusion of layout and semantics.}
Although manually restricting the influence region of each semantic component with the layout control map can manipulate the position of synthesized objects, the synthesis of specific objects fails to benefit from the global textual context.
To appropriately incorporate our layout representations into the image synthesis process and synthesize high-quality images with desired layouts, we propose an adaptive layout-semantic fusion module (PLACE). 
As depicted in Fig.~\ref{fig:overview} (b), in each fusion module, the time embedding is fed into a linear layer to predict an adaptive fusion parameter $\alpha$, which is then used to integrate the layout control map $L$ and the cross attention map $A^{ca}$ to produce the adaptive fusion map $F$ and final output feature $O$:
\begin{equation}
F = \alpha \left(\text{softmax}(L\odot A^{ca})\right) + (1-\alpha) A^{ca}, \  O \!=\! FV.
\end{equation}
By adopting a timestep adaptive parameter $\alpha$ as the weight to fuse layout and semantic features, the global interactions between image tokens and text tokens in the Stable Diffusion are maintained. 
The interactive mechanism allows each image token to access contextual information from a larger set of textual tokens.
Such an adaptive integration of layout and semantics not only helps control the layout of the synthesized image but also facilitates the synthesis of high-quality details.
The trend of learned adaptive $\alpha$ changing over timestep validates the effectiveness of our approach.
As shown in Fig.~\ref{fig:fig4}~(a), the relatively large value of $\alpha$ during the early stages of sampling indicates the crucial role of the layout control map in determining the initial layout.
However, in later stages, the adaptive $\alpha$ gradually decreases, suggesting that the influence of the layout control map diminishes as the process proceeds. 
This enables the image token could actively interact with global textual tokens, thereby synthesizing more realistic details and high-quality results.
The Fig.~\ref{fig:fig4}~(c) shows the variations of the predicted $\hat{x}_0$ during the image synthesis process for both fixed $\alpha$ ($\alpha=1$) and adaptive $\alpha$. 
One can see that in the early sampling phase, the layout of $\hat{x}_0$ is determined, while in the later stages, the model mainly synthesizes realistic details. 
Moreover, compared to the fixed $\alpha=1$ case, adaptive fusion allows the image token to extract information from a greater set of text tokens, enabling synthesizing images with richer and more realistic details, such as the ‘pedestrian crossing’ in the ‘road’ shown in the figure. 
This is also corroborated in Fig.~\ref{fig:fig4}(b), where the text token `sidewalk' not only influences its corresponding semantic region but also affects other contextual regions, such as the `road' class. 
However, this kind of interaction could not be observed under the fixed $\alpha$ condition.

\subsection{Learning Objective}

\begin{figure}[t]
\vspace{-10pt}
\centering
\includegraphics[width=1.0\linewidth]{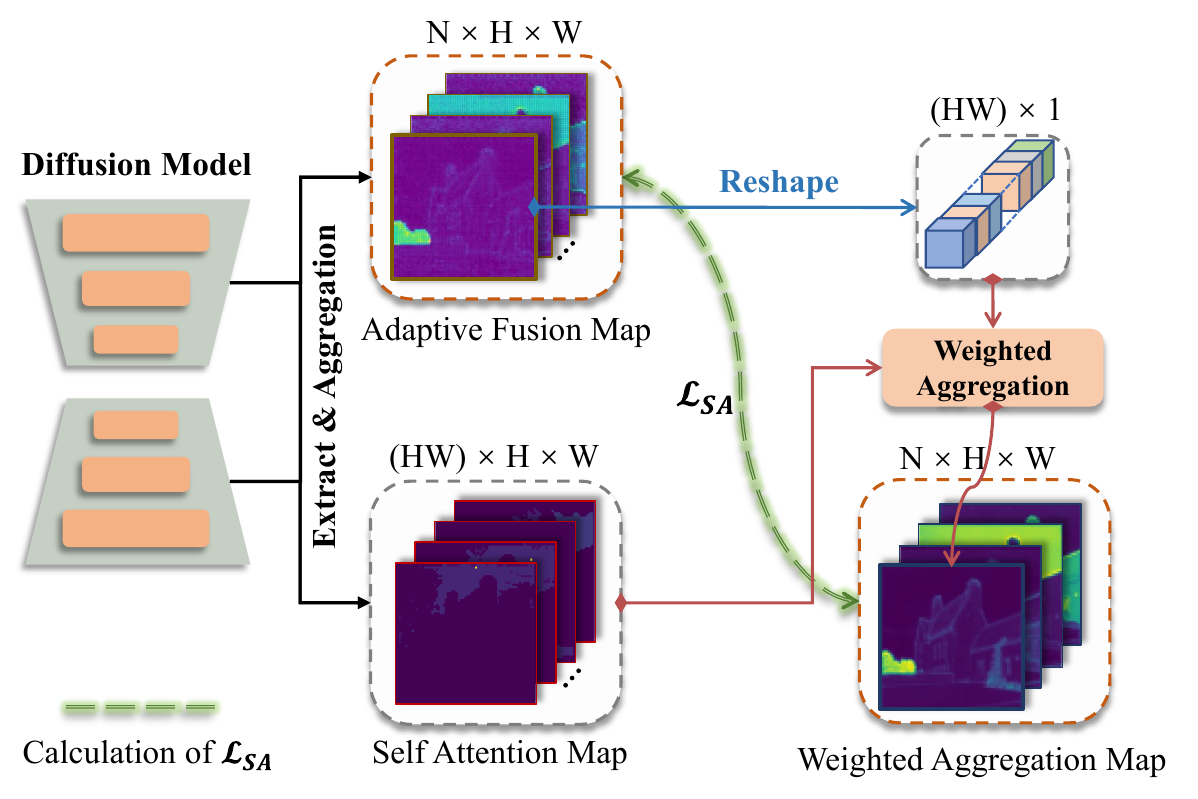}
\vspace{-20pt}
\caption{Calculation of the Semantic Alignment loss.}
\label{fig:fig_saloss}
\vspace{-20pt}
\end{figure}

During the fine-tuning stage, in addition to the original text-to-image denoising loss, we also introduce a semantic alignment~(SA) loss and a layout-free prior preservation~(LFP) loss to facilitate the learning.

\noindent \textbf{Semantic Alignment Loss.} 
To further enhance the layout alignment of synthesized images, we propose the semantic alignment loss $\mathcal{L}_{SA}$. 
As illustrated in Fig.~\ref{fig:fig_saloss}, we first utilize the adaptive fusion map $F\in \mathbb{R}^{N\times H\times W}$ as weights to aggregate the self-attention map $A^{sa} \in \mathbb{R}^{(HW)\times H\times W} $, resulting in weighted aggregation maps $W\in \mathbb{R}^{N\times H \times W}$. 
We then aim to minimize the difference between them and the original adaptive fusion maps, which can be formulated as:
\vspace{-8pt}
\begin{equation}
\vspace{-5pt}
\label{eq:sa}
\begin{aligned}
    W_i = \sum_{j} Reshape(F_i)_j \cdot A^{sa}_j, \\
    \mathcal{L}_{SA} = \sum_{i}||W_i - F_i||^2,
\end{aligned}
\vspace{-5pt}
\end{equation}
where $Reshape(\cdot)$ denotes the flatten operation and $Reshape(F_i)\in \mathbb{R}^{(HW)\times 1}$. 
%
% The semantic alignment loss
%
$\mathcal{L}_{SA}$ effectively encourages image tokens to interact more with the same and related semantic regions in the self-attention module, thereby further improving the layout alignment of the generated images.

\noindent\textbf{Layout-Free Prior Preservation Loss.} 
Due to the limited scale of the fine-tuning dataset, the model inevitably suffers from loss of semantic priors, resulting in suboptimal performance of semantic consistency and visual quality. 
Enlarging the scale of the fine-tuning dataset is one possible way to address this issue. 
However, obtaining a substantial number of real images annotated with semantic masks is non-trivial.

We introduce a Layout-Free Prior Preservation (LFP) loss to alleviate this issue. 
It relies solely on text-image data pairs to help preserve the prior knowledge of the pre-trained model, which is relatively more accessible. 
During each fine-tuning iteration, in addition to sampling regular paired training data with semantic mask annotations $<z_t, S, y, t>$, we also extract an additional set of text-image data pairs $<z_t', y', t'>$ from the Layout Free~(LF) dataset to feed into the network, as shown in Fig.~\ref{fig:overview} (a). 
Due to the absence of semantic masks $S'$, we explicitly set the adaptive fusion parameter $\alpha$ to $0$ when synthesizing the image. 
The original denoising loss $\mathcal{L}_{LDM}$ and our LFP loss $\mathcal{L}_{LFP}$ can be computed as follows:
\begin{equation}
\label{eq:LDM}
\begin{aligned}
\mathcal{L}_{LDM}:=\mathbb{E}_{\mathcal{E}(x),y,\epsilon, t,S}[||\epsilon - \epsilon_{\theta}(z_t,t,S,\tau_{\theta}(y))||^2_2,
\end{aligned}
\end{equation}
\begin{equation}
\label{eq:LFP}
\begin{aligned}
\mathcal{L}_{LFP}:=\mathbb{E}_{\mathcal{E}(x),y',\epsilon', t'}||\epsilon' - \epsilon_{\theta,\alpha=0}(z_t',t',\tau_{\theta}(y'))||^2_2.
\end{aligned}
\end{equation}
We collect approximately 300k text-image pairs from OpenaImages~\cite{kuznetsova2020open} and Laion-5b~\cite{schuhmann2022laion} datasets as the Layout Free dataset.
More details about the implementation of the LFP loss can be found in supplementary materials.

Through employing the LFP loss, semantic concepts present in the pre-trained model are better preserved in the fine-tuning process, even without the involvement of semantic masks. Experimental results demonstrate that our model can generate diverse images and exhibit improved performance in visual quality and semantic consistency.

The optimization objective can be summarized by Eq.~\ref{eq:sa}, Eq.~\ref{eq:LDM}, and Eq.~\ref{eq:LFP} as follows:
\begin{equation}
    \mathcal{L} = \mathcal{L}_{LDM} + \lambda_{1}\mathcal{L}_{SA} + \lambda_{2}\mathcal{L}_{LFP},
\end{equation}
where $\lambda_1$ and $\lambda_2$ are the weight coefficients. And they are set to 1 as default.

\section{Experiments}

\begin{figure*}[t]
\vspace{-1pt}
\centering
\includegraphics[width=0.88\linewidth]{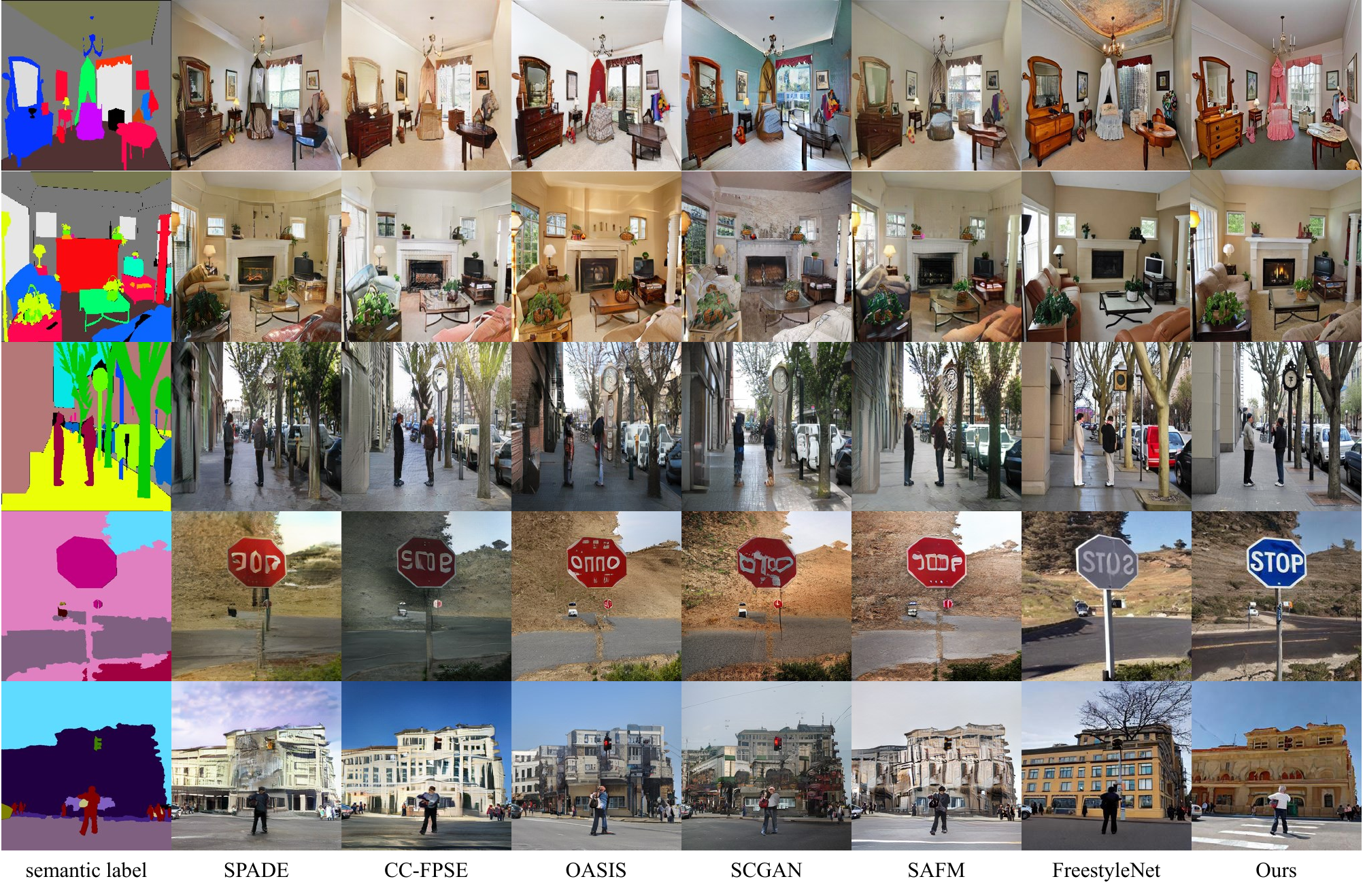}
\vspace{-13pt}
\caption{Visual comparisons on ADE20K ($1st \sim 3rd$ rows) and  COCO-Stuff ($4th \sim 5th$ rows).}
\label{fig:fig5}
\vspace{-15pt}
\end{figure*}

\subsection{Experimental Details}

\noindent\textbf{Datasets.} 
We conduct our experiments on two challenging datasets, namely ADE20K~\cite{zhou2017scene} and COCO-Stuff~\cite{caesar2018coco}. 
ADE20K consists of 150 semantic categories. 
It has 20,210 images for training and 2,000 images available for validation. 
COCO-Stuff contains 182 semantic categories covering diverse scenes. 
It comprises 118,287 training images and 5,000 validation images. 
During training, both the images and semantic maps are resized to $512\times 512$. 
All the semantic classes present in the image are joined together with spaces to form the input textual prompt.

\noindent\textbf{Implementation Details.} We utilize the pre-trained V1-4 Stable Diffusion model~\cite{rombach2022high} as the initialization weights and fine-tune it with a learning rate of $5\times10^{-6}$. All experiments are conducted on a server with 4 NVIDIA V100 32G GPUs. We fine-tuned for about 300k iterations with a batch size of 4. During sampling, we employ 50 PLMS~\cite{liu2022pseudo} sampling steps with a classifier-free guidance~\cite{ho2022classifier} scale of 2.

\noindent\textbf{Evaluation Metrics.} 
Following prior works on semantic image synthesis~\cite{xue2023freestyle}, we quantitatively evaluate the results of in-distribution synthesis with Fréchet Inception Distance (FID)~\cite{heusel2017gans} and the mean Intersection over Union (mIoU). 
FID assesses the visual quality of generated images, while the mIoU measures semantic and layout consistency.
Besides, benefiting from the superior prior of the pre-trained model, our method also shows the capability for out-of-distribution synthesis.
We evaluate this ability from three perspectives, namely new object, new style, and new attribute. 
For new object synthesis, we employ the model fine-tuned on ADE20K to synthesize semantic categories that exclusively appear in COCO-Stuff (\ie, categories not contained in the ADE20K).
FID and mIoU are adopted to assess the quality and consistency of the results. 
Regarding new style and new attribute synthesis, we synthesize 260 images with 8 new global styles and 6 specific object attributes with the same model. 
We measure the consistency of the synthesized results with specific styles or attributes using CLIP~\cite{radford2021learning} text-image similarity (\ie, text alignment). More details can be found in supplementary materials.

\begin{table}
    \vspace{-2pt}

	%\begin{center}
		%\resizebox{\textwidth}{18mm}{
		\small
		\renewcommand\arraystretch{1.}
		\setlength{\tabcolsep}{2.5mm}
		{
			\begin{tabular}{c c c c c c}
				%\hline
				\toprule[1.25pt]
				& \multicolumn{2}{c}{\textbf{ADE20K}} & \multicolumn{2}{c}{\textbf{COCO-Stuff}}\\
				\multirow{-2}{*}{\makecell[c]{\textbf{Methods}}} & \textbf{mIoU $\uparrow$} & \textbf{FID $\downarrow$} & \textbf{mIoU $\uparrow$} & \textbf{FID $\downarrow$} \\
				\hline
				pix2pixHD~\cite{wang2018high}   & 20.3  &  81.8  & 14.6  & 111.5\\
				SPADE~\cite{park2019semantic}   & 38.5 & 33.9 & 37.4  & 22.6  \\
				CC-FPSE~\cite{liu2019learning} & 43.7 &  31.7 & 41.6 &  19.2  \\
				LGGAN~\cite{tang2020local} & 41.6 &  31.6 & N/A  & N/A  \\
				OASIS~\cite{sushko2020you} & 48.3 & 28.3 & 44.1 & 17.0 \\
				SC-GAN~\cite{wang2021image}  &45.2 & 29.3 & 42.0 &18.1  \\
				SAFM~\cite{lv2022semantic} & 50.1 & 32.8 & 43.3 & 24.6 \\
                    RESAIL~\cite{shi2022retrieval} & 49.3 & 30.2 & \underline{44.7} & 18.3 \\
                    ECGAN~\cite{tang2023edge} & \underline{50.6} & 25.8 & \textbf{46.3} & 15.7 \\
                    \hline 

                    SDM~\cite{wang2022semantic} & 39.2 & 27.5 & 40.2 & 15.9 \\
                    PITI~\cite{wang2022pretraining} & 29.4 & 27.9 & 34.1 & 16.1 \\
                    ControlNet~\cite{zhang2023adding} & 36.9 & 31.2 & N/A& N/A\\
                    T2I-Adapter~\cite{mou2023t2i} & N/A& N/A& 20.7 & 16.8 \\
				FreestyleNet~\cite{xue2023freestyle} & 41.9 &\underline{25.0} & 40.7& \underline{14.4} \\
                    Ours & \textbf{50.7} & \textbf{22.3} & 42.6 & \textbf{14.0} \\
				\bottomrule[1.25pt]
			\end{tabular}
		}
	%\end{center}
	\vspace{-8pt}
	\caption{Quantitative comparison on the ADE20K and COCO-Stuff. The upper row shows the results of GAN-based methods, while the lower row displays the scores of those based on diffusion models. $\uparrow$ ($\downarrow$) indicates higher (lower) is better.}
	\label{tab:quan}
	\vspace{-6pt}
\end{table}

\begin{table}
		\vspace{-2pt}
		\begin{center}
			\small
			\renewcommand\arraystretch{1.}
			\setlength{\tabcolsep}{1.0mm}
			{
				\begin{tabular}{c c c c c c}
				%\hline
				\toprule[1.5pt]
				& \multicolumn{2}{c}{\textbf{New Obj.}} & \multicolumn{1}{c}{\textbf{New Sty.}} & \multicolumn{1}{c}{\textbf{New Attri.}} \\
				\multirow{-2}{*}{\makecell[c]{\textbf{Methods}}} & \scriptsize{\textbf{mIoU ($\uparrow$)}} & \scriptsize{\textbf{FID ($\downarrow$)}} & \textbf{\scriptsize{Text-Alignment ($\uparrow$)}} & \textbf{\scriptsize{Text-Alignment ($\uparrow$)}} \\
					\hline
                        ControlNet & 18.2 & 27.4 & 0.274 & 0.284 \\
					FreestyleNet & 24.6  & 20.4 & 0.260 & 0.269   \\
					Ours&  \textbf{33.0} & \textbf{18.1} & \textbf{0.279} &  \textbf{0.290} \\
					%\hline
                    \bottomrule[1.5pt]
				\end{tabular}
			}
		\end{center}
		\vspace{-14pt}
		\caption{Comparison in out-of-distribution synthesis. $\uparrow$ ($\downarrow$) indicates higher (lower) is better.}
		\label{tab:openset}
		\vspace{-14pt}
	\end{table}

\subsection{Evaluation of In-distribution Synthesis}
\vspace{-2pt}
\textbf{Quantitative comparisons.} 
Table~\ref{tab:quan} reports the FID and mIoU performance of our approach compared to other competing methods. 
The upper rows present the results of GAN-based methods, while the lower rows display the scores of methods based on pre-trained text-to-image models. 
As shown, our method achieves FID scores of 22.3 and 14.0 on the ADE20K and COCO-Stuff datasets, respectively, which are 2.7 and 0.4 lower than the second-best scores. 
In terms of alignment, our method obtains results comparable to the state-of-the-art. 
On the ADE20K, our mIoU score reaches 50.7, while on the COCO-Stuff, our score is 42.6. 
The quantitative results indicate that our method not only achieves comparable performance in semantic and layout consistency to the current state-of-the-art works but also attains the highest image quality scores.
The reliable layout representation allows our approach to demonstrate enhanced consistency in layout details compared to other methods based on pre-trained text-to-image models, on par with the most advanced GAN-based methods.
Moreover, the efficient interaction between layout and semantic features in the adaptive layout-semantic fusion plays a vital role in synthesizing high-quality images. 

\noindent\textbf{Qualitative comparisons.}
Fig.~\ref{fig:fig5} illustrates the qualitative comparisons on the ADE20K and COCO-Stuff, from which the following observations can be made: 
(1) Our method produces synthesis results that exhibit higher fidelity to the semantic layout. 
For example, the `clock' in the $3rd$ row demonstrates improved alignment with the semantic layout.
(2) The images synthesized by our method demonstrate more realistic details. 
Notable examples include the `bed' in the $1st$ row, the `table' in the $2nd$ row, and the `pedestrian crossing' in the $5th$ row. 
(3) Our method effectively preserves and utilizes the priors in the pre-trained model. 
An example can be seen in the $4th$ row with the preservation of the signage.
More qualitative results can be found in supplementary materials.

\vspace{-3pt}
\subsection{Evaluation of Out-distribution Synthesis}
\begin{figure}
\vspace{-4pt}
\centering
\includegraphics[width=0.92\linewidth]{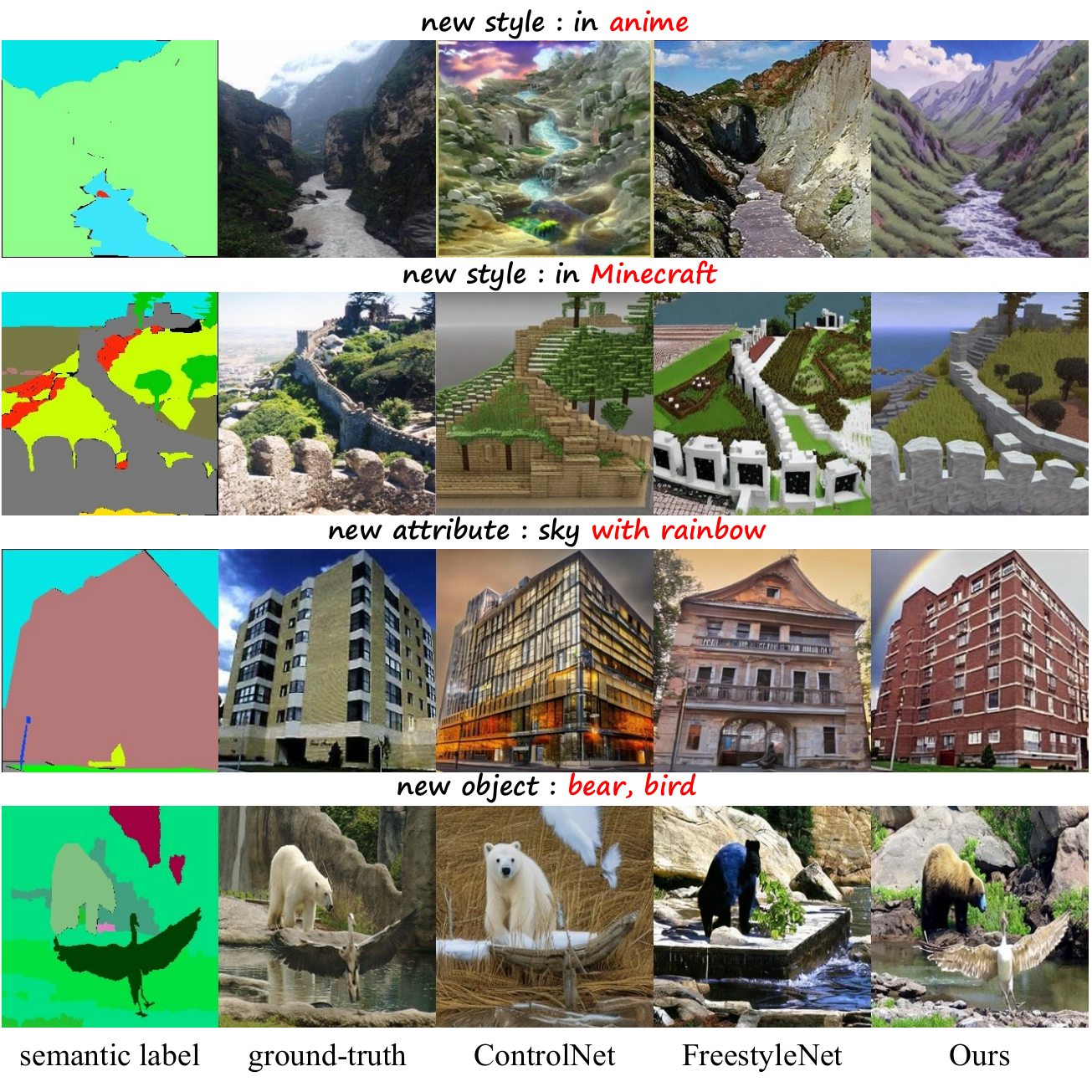}
\vspace{-11pt}
\caption{Visual comparisons for out-of-distribution synthesis. %Zoom in for details.
}
\label{fig:fig6}
\vspace{-20pt}
\end{figure}

\vspace{-3pt}
Table~\ref{tab:openset} and Fig.~\ref{fig:fig6} present the quantitative and qualitative comparisons of out-of-distribution synthesis results respectively. 
The evaluation of out-of-distribution synthesis comprises three aspects, namely new object, new style, and new attribute. As shown, our method achieves superior quantitative scores in all three aspects compared to both ControlNet and FreestyleNet. 
Especially in the new object category, our method achieves a significant mIoU improvement of 8.4 compared to FreestyleNet. 
From the visual comparison, it is evident that our method synthesizes out-of-distribution images with not only better semantic consistency with the given conditions (\ie, new semantic, new style, and new attribute) but also maintains good performance in terms of layout alignment. 
For instance, in Fig.~\ref{fig:fig6}, the `anime' style in the $1st$ row, the `rainbow' in the $3rd$ row, and the `bird' in the $4th$ row are all faithfully consistent with the provided conditions. 
The `bird' and the `bear' in the $4th$ row demonstrate strong layout alignment. 
More results can be found in supplementary materials.

\vspace{-3pt}
\subsection{Ablation Study}
\vspace{-3pt}
We conduct the ablation study with variant models fine-tuned on the ADE20K dataset to validate the effectiveness of our method.
Our baseline model employs a simple nearest resized semantic map to determine the region of influence for each text token based on given semantic maps. 
It does not involve adaptive fusion and is fine-tuned only with denoising loss. 
Besides, we utilize a more comprehensive vocabulary, wherein a greater number of synonyms are employed to represent the same semantic category.

\begin{figure}
\vspace{-1pt}
\centering
\includegraphics[width=1.\linewidth]{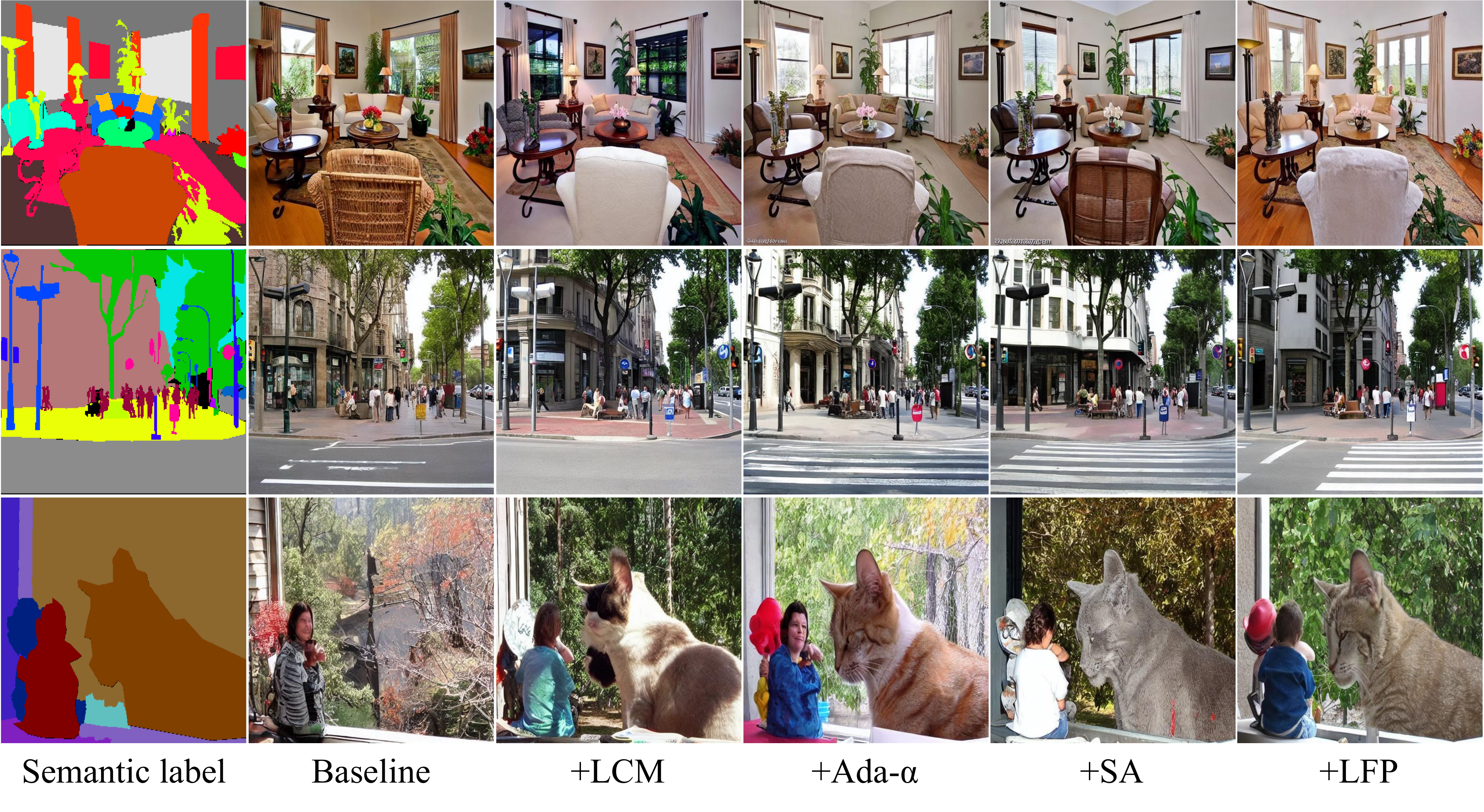}
\vspace{-24pt}
\caption{Visual comparisons of different variants.}
\label{fig:fig7}
\vspace{-8pt}
\end{figure}

The comparisons of quantitative and qualitative results are presented in Table~\ref{tab:ablation} and Fig.~\ref{fig:fig7}, respectively. 
In Table~\ref{tab:ablation}, LCM denotes the Layout Control Map. 
Ada-$\alpha$ indicates the usage of the timestep-adaptive parameter during fusion, SA represents Semantic Alignment loss, and LFP refers to Layout-Free Prior Preservation loss. 
The \checkmark indicates the adoption of the corresponding module or strategy during the experiments. We compared the FID and mIoU scores of ADE20K and the new object classes~(denoted by ‘New Obj.’) from COCO-Stuff. 
More qualitative results can be found in supplementary materials.

\noindent\textbf{Layout control map.} 
From (1) and (2) in Table~\ref{tab:ablation}, the layout control map significantly improves the mIoU scores, increasing from 43.5 to 48.6 for in-distribution synthesis and from 25.3 to 28.5 for out-of-distribution synthesis. 
Besides, from Fig.~\ref{fig:fig7}, with the layout control map, our method can generate images that adhere closely to given layouts (\eg, the `table' and `plants' in $1st$ row, and the `street light' in $2nd$ row), demonstrating its effectiveness.

\noindent \textbf{Adaptive $\alpha$ for fusion.}
Referring to the (1) and (3) as well as the (2) and (5) in Table~\ref{tab:ablation}, with the adaptive layout-semantic fusion, the FID scores on the ADE20K decrease by 1.2 and 0.7, respectively.
It is evident that the adaptive fusion enhances the quality of the synthesized images. 
In Fig.~\ref{fig:fig7}, the results using adaptive fusion exhibit more realistic details (\eg, `window' and `road' in $1st$ and $2nd$ rows).

%\vspace{-1em}
\noindent \textbf{Semantic Alignment loss.} 
Both the (1) and (4) along with the (5) and (6) in Table~\ref{tab:ablation} indicate that the semantic alignment loss contributes to the consistency of the layout. 
The mIoU scores increase by 3.2 and 0.8 on the ADE20K, individually. 
As shown in Fig.~\ref{fig:fig7}, the alignment loss also helps synthesize more realistic instances (\eg, $3rd$ row).

\vspace{-1pt}
\noindent\textbf{Layout Free Prior Preservation loss.} 
\vspace{-1pt}
The LFP loss better preserves the semantic priors in the pre-trained model, resulting in improved performance for visual quality and semantic consistency. 
The (5) and (7) in Table~\ref{tab:ablation} show that the LFP loss leads to a 3.4 increase in the mIoU score and a 1.2 decrease in the FID score for new object synthesis. From experiments (6) and (8), the mIoU and FID scores of synthesized new objects improved by 3.1 and 1.3, respectively. With the LFP loss, the $3rd$ row in Fig.~\ref{fig:fig7} presents a more realistic `cat', even though this category does not appear in the training dataset.

\begin{table}
		\vspace{-2pt}
		\begin{center}
			\small
			\renewcommand\arraystretch{1}
			\setlength{\tabcolsep}{0.8mm}
			{
				\begin{tabular}{c c c c c  c c c c}
				\toprule[1.25pt]
       & \multicolumn{4}{c}{\textbf{Methods}} & \multicolumn{2}{c}{\textbf{ADE20K}} & \multicolumn{2}{c}{\textbf{New Obj.}} \\
     & \textbf{LCM} & \textbf{Ada-$\alpha$} & \textbf{SA} & \textbf{LFP}
                & \textbf{mIoU $\uparrow$} & \textbf{FID $\downarrow$} 
                & \textbf{mIoU* $\uparrow$} & \textbf{FID $\downarrow$} \\
					\hline
				(1)& & & & & 43.5 & 24.2 & 25.3 & 20.2  \\
					(2)& \checkmark & & & & 48.6 & 23.4 & 28.5 & 19.8\\
                    (3)& & \checkmark &  & &46.2&23.0  &26.5  & 19.4 \\
                    (4) & & & \checkmark & & 46.7 & 23.9 & 27.1 & 19.9 \\
				(5)& \checkmark & \checkmark & & & 50.1 & \underline{22.7} & 29.4 & \underline{19.3}  \\ %latest_76_78295
                    (6) & \checkmark & \checkmark& \checkmark & & \textbf{50.9} & 22.8 & 29.9& 19.4 \\
					(7)& \checkmark & \checkmark & & \checkmark & 49.8 & \textbf{22.3} & \underline{32.8} & \textbf{18.1}  \\ 
                    (8)& \checkmark & \checkmark & \checkmark & \checkmark & \underline{50.7} & \textbf{22.3} & \textbf{33.0} & \textbf{18.1} \\
					\bottomrule[1.25pt]
				\end{tabular}
			}
		\end{center}
		\vspace{-15pt}
		\caption{Quantitative comparison of five variants on Ade20K in the ablation study. The mIoU* denotes the mIoU scores of semantic classes that are exclusively present in the COCO-Stuff dataset.}
		\label{tab:ablation}
		\vspace{-20pt}
	\end{table}

\vspace{-10pt}
\section{Conclusion}
\vspace{-6pt}

In this paper, we first present a novel layout control map for reliable representations of layout features. 
We further combine the semantic and layout features adaptively, resulting in the synthesis of high-quality images that are faithfully aligned with given semantic layouts. 
Additionally, we propose a semantic alignment loss to facilitate the layout alignment and a layout-free prior preservation loss to maintain semantic priors of pre-trained models for fine-tuning.
Extensive quantitative and qualitative results demonstrate that PLACE exhibits remarkable visual quality, semantic consistency, and layout alignment for both the in-distribution and out-of-distribution semantic image synthesis. \par

% \section{Final copy}

% You must include your signed IEEE copyright release form when you submit your finished paper.
% We MUST have this form before your paper can be published in the proceedings.

% Please direct any questions to the production editor in charge of these proceedings at the IEEE Computer Society Press:
% \url{https://www.computer.org/about/contact}.
{
    \small
    \bibliographystyle{ieeenat_fullname}
    \bibliography{main}

\begin{thebibliography}{50}
\providecommand{\natexlab}[1]{#1}
\providecommand{\url}[1]{\texttt{#1}}
\expandafter\ifx\csname urlstyle\endcsname\relax
  \providecommand{\doi}[1]{doi: #1}\else
  \providecommand{\doi}{doi: \begingroup \urlstyle{rm}\Url}\fi

\bibitem[Avrahami et~al.(2023)Avrahami, Hayes, Gafni, Gupta, Taigman, Parikh, Lischinski, Fried, and Yin]{avrahami2023spatext}
Omri Avrahami, Thomas Hayes, Oran Gafni, Sonal Gupta, Yaniv Taigman, Devi Parikh, Dani Lischinski, Ohad Fried, and Xi Yin.
\newblock Spatext: Spatio-textual representation for controllable image generation.
\newblock In \emph{Proceedings of the IEEE/CVF Conference on Computer Vision and Pattern Recognition}, pages 18370--18380, 2023.

\bibitem[Balaji et~al.(2022)Balaji, Nah, Huang, Vahdat, Song, Kreis, Aittala, Aila, Laine, Catanzaro, et~al.]{balaji2022ediffi}
Yogesh Balaji, Seungjun Nah, Xun Huang, Arash Vahdat, Jiaming Song, Karsten Kreis, Miika Aittala, Timo Aila, Samuli Laine, Bryan Catanzaro, et~al.
\newblock ediffi: Text-to-image diffusion models with an ensemble of expert denoisers.
\newblock \emph{arXiv preprint arXiv:2211.01324}, 2022.

\bibitem[Caesar et~al.(2018)Caesar, Uijlings, and Ferrari]{caesar2018coco}
Holger Caesar, Jasper Uijlings, and Vittorio Ferrari.
\newblock Coco-stuff: Thing and stuff classes in context.
\newblock In \emph{Proceedings of the IEEE conference on computer vision and pattern recognition}, pages 1209--1218, 2018.

\bibitem[Chen et~al.(2023)Chen, Laina, and Vedaldi]{chen2023training}
Minghao Chen, Iro Laina, and Andrea Vedaldi.
\newblock Training-free layout control with cross-attention guidance.
\newblock \emph{arXiv preprint arXiv:2304.03373}, 2023.

\bibitem[Chen and Koltun(2017)]{chen2017photographic}
Qifeng Chen and Vladlen Koltun.
\newblock Photographic image synthesis with cascaded refinement networks.
\newblock In \emph{Proceedings of the IEEE international conference on computer vision}, pages 1511--1520, 2017.

\bibitem[Goel et~al.(2023)Goel, Peruzzo, Jiang, Xu, Sebe, Darrell, Wang, and Shi]{goel2023pair}
Vidit Goel, Elia Peruzzo, Yifan Jiang, Dejia Xu, Nicu Sebe, Trevor Darrell, Zhangyang Wang, and Humphrey Shi.
\newblock Pair-diffusion: Object-level image editing with structure-and-appearance paired diffusion models.
\newblock \emph{arXiv preprint arXiv:2303.17546}, 2023.

\bibitem[Goodfellow et~al.(2014)Goodfellow, Pouget-Abadie, Mirza, Xu, Warde-Farley, Ozair, Courville, and Bengio]{goodfellow2014generative}
Ian Goodfellow, Jean Pouget-Abadie, Mehdi Mirza, Bing Xu, David Warde-Farley, Sherjil Ozair, Aaron Courville, and Yoshua Bengio.
\newblock Generative adversarial nets.
\newblock \emph{Advances in neural information processing systems}, 27, 2014.

\bibitem[Hao et~al.(2023)Hao, Han, Zhao, and Wong]{hao2023vico}
Shaozhe Hao, Kai Han, Shihao Zhao, and Kwan-Yee~K Wong.
\newblock Vico: Detail-preserving visual condition for personalized text-to-image generation.
\newblock \emph{arXiv preprint arXiv:2306.00971}, 2023.

\bibitem[Heusel et~al.(2017)Heusel, Ramsauer, Unterthiner, Nessler, and Hochreiter]{heusel2017gans}
Martin Heusel, Hubert Ramsauer, Thomas Unterthiner, Bernhard Nessler, and Sepp Hochreiter.
\newblock Gans trained by a two time-scale update rule converge to a local nash equilibrium.
\newblock \emph{Advances in neural information processing systems}, 30, 2017.

\bibitem[Ho and Salimans(2022)]{ho2022classifier}
Jonathan Ho and Tim Salimans.
\newblock Classifier-free diffusion guidance.
\newblock \emph{arXiv preprint arXiv:2207.12598}, 2022.

\bibitem[Ho et~al.(2020)Ho, Jain, and Abbeel]{ho2020denoising}
Jonathan Ho, Ajay Jain, and Pieter Abbeel.
\newblock Denoising diffusion probabilistic models.
\newblock \emph{Advances in neural information processing systems}, 33:\penalty0 6840--6851, 2020.

\bibitem[Isola et~al.(2017)Isola, Zhu, Zhou, and Efros]{isola2017image}
Phillip Isola, Jun-Yan Zhu, Tinghui Zhou, and Alexei~A Efros.
\newblock Image-to-image translation with conditional adversarial networks.
\newblock In \emph{Proceedings of the IEEE conference on computer vision and pattern recognition}, pages 1125--1134, 2017.

\bibitem[Kim et~al.(2023)Kim, Lee, Kim, Ha, and Zhu]{kim2023dense}
Yunji Kim, Jiyoung Lee, Jin-Hwa Kim, Jung-Woo Ha, and Jun-Yan Zhu.
\newblock Dense text-to-image generation with attention modulation.
\newblock In \emph{Proceedings of the IEEE/CVF International Conference on Computer Vision}, pages 7701--7711, 2023.

\bibitem[Kuznetsova et~al.(2020)Kuznetsova, Rom, Alldrin, Uijlings, Krasin, Pont-Tuset, Kamali, Popov, Malloci, Kolesnikov, et~al.]{kuznetsova2020open}
Alina Kuznetsova, Hassan Rom, Neil Alldrin, Jasper Uijlings, Ivan Krasin, Jordi Pont-Tuset, Shahab Kamali, Stefan Popov, Matteo Malloci, Alexander Kolesnikov, et~al.
\newblock The open images dataset v4: Unified image classification, object detection, and visual relationship detection at scale.
\newblock \emph{International Journal of Computer Vision}, 128\penalty0 (7):\penalty0 1956--1981, 2020.

\bibitem[Liu et~al.(2022)Liu, Ren, Lin, and Zhao]{liu2022pseudo}
Luping Liu, Yi Ren, Zhijie Lin, and Zhou Zhao.
\newblock Pseudo numerical methods for diffusion models on manifolds.
\newblock \emph{arXiv preprint arXiv:2202.09778}, 2022.

\bibitem[Liu et~al.(2019)Liu, Yin, Shao, Wang, and Li]{liu2019learning}
Xihui Liu, Guojun Yin, Jing Shao, Xiaogang Wang, and Hongsheng Li.
\newblock Learning to predict layout-to-image conditional convolutions for semantic image synthesis.
\newblock \emph{arXiv preprint arXiv:1910.06809}, 2019.

\bibitem[Luo et~al.(2022)Luo, Yang, Wang, Long, and Zhang]{luo2022context}
Wuyang Luo, Su Yang, Hong Wang, Bo Long, and Weishan Zhang.
\newblock Context-consistent semantic image editing with style-preserved modulation.
\newblock In \emph{European Conference on Computer Vision}, pages 561--578. Springer, 2022.

\bibitem[Luo et~al.(2023)Luo, Yang, Zhang, and Zhang]{luo2023siedob}
Wuyang Luo, Su Yang, Xinjian Zhang, and Weishan Zhang.
\newblock Siedob: Semantic image editing by disentangling object and background.
\newblock In \emph{Proceedings of the IEEE/CVF Conference on Computer Vision and Pattern Recognition}, pages 1868--1878, 2023.

\bibitem[Lv et~al.(2022)Lv, Li, Niu, Cao, and Zuo]{lv2022semantic}
Zhengyao Lv, Xiaoming Li, Zhenxing Niu, Bing Cao, and Wangmeng Zuo.
\newblock Semantic-shape adaptive feature modulation for semantic image synthesis.
\newblock In \emph{Proceedings of the IEEE/CVF Conference on Computer Vision and Pattern Recognition}, pages 11214--11223, 2022.

\bibitem[Mou et~al.(2023)Mou, Wang, Xie, Zhang, Qi, Shan, and Qie]{mou2023t2i}
Chong Mou, Xintao Wang, Liangbin Xie, Jian Zhang, Zhongang Qi, Ying Shan, and Xiaohu Qie.
\newblock T2i-adapter: Learning adapters to dig out more controllable ability for text-to-image diffusion models.
\newblock \emph{arXiv preprint arXiv:2302.08453}, 2023.

\bibitem[Nichol et~al.(2021)Nichol, Dhariwal, Ramesh, Shyam, Mishkin, McGrew, Sutskever, and Chen]{nichol2021glide}
Alex Nichol, Prafulla Dhariwal, Aditya Ramesh, Pranav Shyam, Pamela Mishkin, Bob McGrew, Ilya Sutskever, and Mark Chen.
\newblock Glide: Towards photorealistic image generation and editing with text-guided diffusion models.
\newblock \emph{arXiv preprint arXiv:2112.10741}, 2021.

\bibitem[Ntavelis et~al.(2020)Ntavelis, Romero, Kastanis, Van~Gool, and Timofte]{ntavelis2020sesame}
Evangelos Ntavelis, Andr{\'e}s Romero, Iason Kastanis, Luc Van~Gool, and Radu Timofte.
\newblock Sesame: Semantic editing of scenes by adding, manipulating or erasing objects.
\newblock In \emph{Computer Vision--ECCV 2020: 16th European Conference, Glasgow, UK, August 23--28, 2020, Proceedings, Part XXII 16}, pages 394--411. Springer, 2020.

\bibitem[Park et~al.(2019)Park, Liu, Wang, and Zhu]{park2019semantic}
Taesung Park, Ming-Yu Liu, Ting-Chun Wang, and Jun-Yan Zhu.
\newblock Semantic image synthesis with spatially-adaptive normalization.
\newblock In \emph{Proceedings of the IEEE/CVF Conference on Computer Vision and Pattern Recognition}, pages 2337--2346, 2019.

\bibitem[Phung et~al.(2023)Phung, Ge, and Huang]{phung2023grounded}
Quynh Phung, Songwei Ge, and Jia-Bin Huang.
\newblock Grounded text-to-image synthesis with attention refocusing.
\newblock \emph{arXiv preprint arXiv:2306.05427}, 2023.

\bibitem[Qi et~al.(2018)Qi, Chen, Jia, and Koltun]{qi2018semi}
Xiaojuan Qi, Qifeng Chen, Jiaya Jia, and Vladlen Koltun.
\newblock Semi-parametric image synthesis.
\newblock In \emph{Proceedings of the IEEE Conference on Computer Vision and Pattern Recognition}, pages 8808--8816, 2018.

\bibitem[Radford et~al.(2021)Radford, Kim, Hallacy, Ramesh, Goh, Agarwal, Sastry, Askell, Mishkin, Clark, et~al.]{radford2021learning}
Alec Radford, Jong~Wook Kim, Chris Hallacy, Aditya Ramesh, Gabriel Goh, Sandhini Agarwal, Girish Sastry, Amanda Askell, Pamela Mishkin, Jack Clark, et~al.
\newblock Learning transferable visual models from natural language supervision.
\newblock In \emph{International conference on machine learning}, pages 8748--8763. PMLR, 2021.

\bibitem[Ramesh et~al.(2022)Ramesh, Dhariwal, Nichol, Chu, and Chen]{ramesh2022hierarchical}
Aditya Ramesh, Prafulla Dhariwal, Alex Nichol, Casey Chu, and Mark Chen.
\newblock Hierarchical text-conditional image generation with clip latents.
\newblock \emph{arXiv preprint arXiv:2204.06125}, 2022.

\bibitem[Rombach et~al.(2022)Rombach, Blattmann, Lorenz, Esser, and Ommer]{rombach2022high}
Robin Rombach, Andreas Blattmann, Dominik Lorenz, Patrick Esser, and Bj{\"o}rn Ommer.
\newblock High-resolution image synthesis with latent diffusion models.
\newblock In \emph{Proceedings of the IEEE/CVF conference on computer vision and pattern recognition}, pages 10684--10695, 2022.

\bibitem[Ruiz et~al.(2023)Ruiz, Li, Jampani, Pritch, Rubinstein, and Aberman]{ruiz2023dreambooth}
Nataniel Ruiz, Yuanzhen Li, Varun Jampani, Yael Pritch, Michael Rubinstein, and Kfir Aberman.
\newblock Dreambooth: Fine tuning text-to-image diffusion models for subject-driven generation.
\newblock In \emph{Proceedings of the IEEE/CVF Conference on Computer Vision and Pattern Recognition}, pages 22500--22510, 2023.

\bibitem[Saharia et~al.(2022)Saharia, Chan, Saxena, Li, Whang, Denton, Ghasemipour, Gontijo~Lopes, Karagol~Ayan, Salimans, et~al.]{saharia2022photorealistic}
Chitwan Saharia, William Chan, Saurabh Saxena, Lala Li, Jay Whang, Emily~L Denton, Kamyar Ghasemipour, Raphael Gontijo~Lopes, Burcu Karagol~Ayan, Tim Salimans, et~al.
\newblock Photorealistic text-to-image diffusion models with deep language understanding.
\newblock \emph{Advances in Neural Information Processing Systems}, 35:\penalty0 36479--36494, 2022.

\bibitem[Schuhmann et~al.(2022)Schuhmann, Beaumont, Vencu, Gordon, Wightman, Cherti, Coombes, Katta, Mullis, Wortsman, et~al.]{schuhmann2022laion}
Christoph Schuhmann, Romain Beaumont, Richard Vencu, Cade Gordon, Ross Wightman, Mehdi Cherti, Theo Coombes, Aarush Katta, Clayton Mullis, Mitchell Wortsman, et~al.
\newblock Laion-5b: An open large-scale dataset for training next generation image-text models.
\newblock \emph{Advances in Neural Information Processing Systems}, 35:\penalty0 25278--25294, 2022.

\bibitem[Shi et~al.(2022)Shi, Liu, Wei, Wu, and Zuo]{shi2022retrieval}
Yupeng Shi, Xiao Liu, Yuxiang Wei, Zhongqin Wu, and Wangmeng Zuo.
\newblock Retrieval-based spatially adaptive normalization for semantic image synthesis.
\newblock In \emph{Proceedings of the IEEE/CVF Conference on Computer Vision and Pattern Recognition}, pages 11224--11233, 2022.

\bibitem[Song et~al.(2020)Song, Sohl-Dickstein, Kingma, Kumar, Ermon, and Poole]{song2020score}
Yang Song, Jascha Sohl-Dickstein, Diederik~P Kingma, Abhishek Kumar, Stefano Ermon, and Ben Poole.
\newblock Score-based generative modeling through stochastic differential equations.
\newblock \emph{arXiv preprint arXiv:2011.13456}, 2020.

\bibitem[Sushko et~al.(2020)Sushko, Sch{\"o}nfeld, Zhang, Gall, Schiele, and Khoreva]{sushko2020you}
Vadim Sushko, Edgar Sch{\"o}nfeld, Dan Zhang, Juergen Gall, Bernt Schiele, and Anna Khoreva.
\newblock You only need adversarial supervision for semantic image synthesis.
\newblock \emph{arXiv preprint arXiv:2012.04781}, 2020.

\bibitem[Tang et~al.(2020)Tang, Xu, Yan, Torr, and Sebe]{tang2020local}
Hao Tang, Dan Xu, Yan Yan, Philip~HS Torr, and Nicu Sebe.
\newblock Local class-specific and global image-level generative adversarial networks for semantic-guided scene generation.
\newblock In \emph{Proceedings of the IEEE/CVF Conference on Computer Vision and Pattern Recognition}, pages 7870--7879, 2020.

\bibitem[Tang et~al.(2023)Tang, Sun, Sebe, and Van~Gool]{tang2023edge}
Hao Tang, Guolei Sun, Nicu Sebe, and Luc Van~Gool.
\newblock Edge guided gans with multi-scale contrastive learning for semantic image synthesis.
\newblock \emph{IEEE Transactions on Pattern Analysis and Machine Intelligence}, 2023.

\bibitem[Wang et~al.(2022{\natexlab{a}})Wang, Zhang, Zhang, Ouyang, Chen, Chen, and Wen]{wang2022pretraining}
Tengfei Wang, Ting Zhang, Bo Zhang, Hao Ouyang, Dong Chen, Qifeng Chen, and Fang Wen.
\newblock Pretraining is all you need for image-to-image translation.
\newblock \emph{arXiv preprint arXiv:2205.12952}, 2022{\natexlab{a}}.

\bibitem[Wang et~al.(2018)Wang, Liu, Zhu, Tao, Kautz, and Catanzaro]{wang2018high}
Ting-Chun Wang, Ming-Yu Liu, Jun-Yan Zhu, Andrew Tao, Jan Kautz, and Bryan Catanzaro.
\newblock High-resolution image synthesis and semantic manipulation with conditional gans.
\newblock In \emph{Proceedings of the IEEE conference on computer vision and pattern recognition}, pages 8798--8807, 2018.

\bibitem[Wang et~al.(2022{\natexlab{b}})Wang, Bao, Zhou, Chen, Chen, Yuan, and Li]{wang2022semantic}
Weilun Wang, Jianmin Bao, Wengang Zhou, Dongdong Chen, Dong Chen, Lu Yuan, and Houqiang Li.
\newblock Semantic image synthesis via diffusion models.
\newblock \emph{arXiv preprint arXiv:2207.00050}, 2022{\natexlab{b}}.

\bibitem[Wang et~al.(2021)Wang, Qi, Chen, Zhang, and Jia]{wang2021image}
Yi Wang, Lu Qi, Ying-Cong Chen, Xiangyu Zhang, and Jiaya Jia.
\newblock Image synthesis via semantic composition.
\newblock \emph{arXiv preprint arXiv:2109.07053}, 2021.

\bibitem[Wei et~al.(2023{\natexlab{a}})Wei, Ji, Wu, Bai, Zhang, and Zuo]{wei2023inferring}
Yuxiang Wei, Zhilong Ji, Xiaohe Wu, Jinfeng Bai, Lei Zhang, and Wangmeng Zuo.
\newblock Inferring and leveraging parts from object shape for improving semantic image synthesis.
\newblock In \emph{Proceedings of the IEEE/CVF Conference on Computer Vision and Pattern Recognition}, pages 11248--11258, 2023{\natexlab{a}}.

\bibitem[Wei et~al.(2023{\natexlab{b}})Wei, Zhang, Ji, Bai, Zhang, and Zuo]{wei2023elite}
Yuxiang Wei, Yabo Zhang, Zhilong Ji, Jinfeng Bai, Lei Zhang, and Wangmeng Zuo.
\newblock Elite: Encoding visual concepts into textual embeddings for customized text-to-image generation.
\newblock \emph{arXiv preprint arXiv:2302.13848}, 2023{\natexlab{b}}.

\bibitem[Xiao et~al.(2023)Xiao, Li, Lv, Wang, and Huang]{xiao2023r}
Jiayu Xiao, Liang Li, Henglei Lv, Shuhui Wang, and Qingming Huang.
\newblock R\&b: Region and boundary aware zero-shot grounded text-to-image generation.
\newblock \emph{arXiv preprint arXiv:2310.08872}, 2023.

\bibitem[Xie et~al.(2023)Xie, Li, Huang, Liu, Zhang, Zheng, and Shou]{xie2023boxdiff}
Jinheng Xie, Yuexiang Li, Yawen Huang, Haozhe Liu, Wentian Zhang, Yefeng Zheng, and Mike~Zheng Shou.
\newblock Boxdiff: Text-to-image synthesis with training-free box-constrained diffusion.
\newblock In \emph{Proceedings of the IEEE/CVF International Conference on Computer Vision}, pages 7452--7461, 2023.

\bibitem[Xue et~al.(2023)Xue, Huang, Sun, Song, and Zhang]{xue2023freestyle}
Han Xue, Zhiwu Huang, Qianru Sun, Li Song, and Wenjun Zhang.
\newblock Freestyle layout-to-image synthesis.
\newblock In \emph{Proceedings of the IEEE/CVF Conference on Computer Vision and Pattern Recognition}, pages 14256--14266, 2023.

\bibitem[Yang et~al.(2023)Yang, Xu, Kang, Shi, and Zhao]{yang2023freemask}
Lihe Yang, Xiaogang Xu, Bingyi Kang, Yinghuan Shi, and Hengshuang Zhao.
\newblock Freemask: Synthetic images with dense annotations make stronger segmentation models.
\newblock \emph{arXiv preprint arXiv:2310.15160}, 2023.

\bibitem[Zhang and Agrawala(2023)]{zhang2023adding}
Lvmin Zhang and Maneesh Agrawala.
\newblock Adding conditional control to text-to-image diffusion models.
\newblock \emph{arXiv preprint arXiv:2302.05543}, 2023.

\bibitem[Zhao et~al.(2024)Zhao, Chen, Chen, Bao, Hao, Yuan, and Wong]{zhao2024uni}
Shihao Zhao, Dongdong Chen, Yen-Chun Chen, Jianmin Bao, Shaozhe Hao, Lu Yuan, and Kwan-Yee~K Wong.
\newblock Uni-controlnet: All-in-one control to text-to-image diffusion models.
\newblock \emph{Advances in Neural Information Processing Systems}, 36, 2024.

\bibitem[Zhou et~al.(2017)Zhou, Zhao, Puig, Fidler, Barriuso, and Torralba]{zhou2017scene}
Bolei Zhou, Hang Zhao, Xavier Puig, Sanja Fidler, Adela Barriuso, and Antonio Torralba.
\newblock Scene parsing through ade20k dataset.
\newblock In \emph{Proceedings of the IEEE conference on computer vision and pattern recognition}, pages 633--641, 2017.

\bibitem[Zhu et~al.(2020)Zhu, Abdal, Qin, and Wonka]{zhu2020sean}
Peihao Zhu, Rameen Abdal, Yipeng Qin, and Peter Wonka.
\newblock Sean: Image synthesis with semantic region-adaptive normalization.
\newblock In \emph{Proceedings of the IEEE/CVF Conference on Computer Vision and Pattern Recognition}, pages 5104--5113, 2020.

\end{thebibliography}
}

\clearpage
\setcounter{page}{1}
\maketitlesupplementary

\renewcommand{\thesection}{\Alph{section}}
\renewcommand{\thetable}{\Alph{table}}
\renewcommand{\thefigure}{\Alph{figure}}
\renewcommand{\theequation}{\Alph{equation}}
\setcounter{section}{0}
\setcounter{figure}{0}
\setcounter{table}{0}
\setcounter{equation}{0}

This supplemental file provides the following materials:

\begin{itemize}
\vspace{0.3em}
\item More details of LFP Loss in Sec.~\ref{sec:more_lfp_loss};
\vspace{0.5em}
\item More ablation studies in Sec.~\ref{sec:more_ablation};
\vspace{0.5em}
\item More qualitative results in Sec.~\ref{sec:more_results};
\vspace{0.5em}
\item Discussion on limitation in Sec.~\ref{sec:limitation}.
\end{itemize}

\section{More Details of LFP Loss}
\label{sec:more_lfp_loss}

\begin{figure}
\centering
\includegraphics[width=1.0\linewidth]{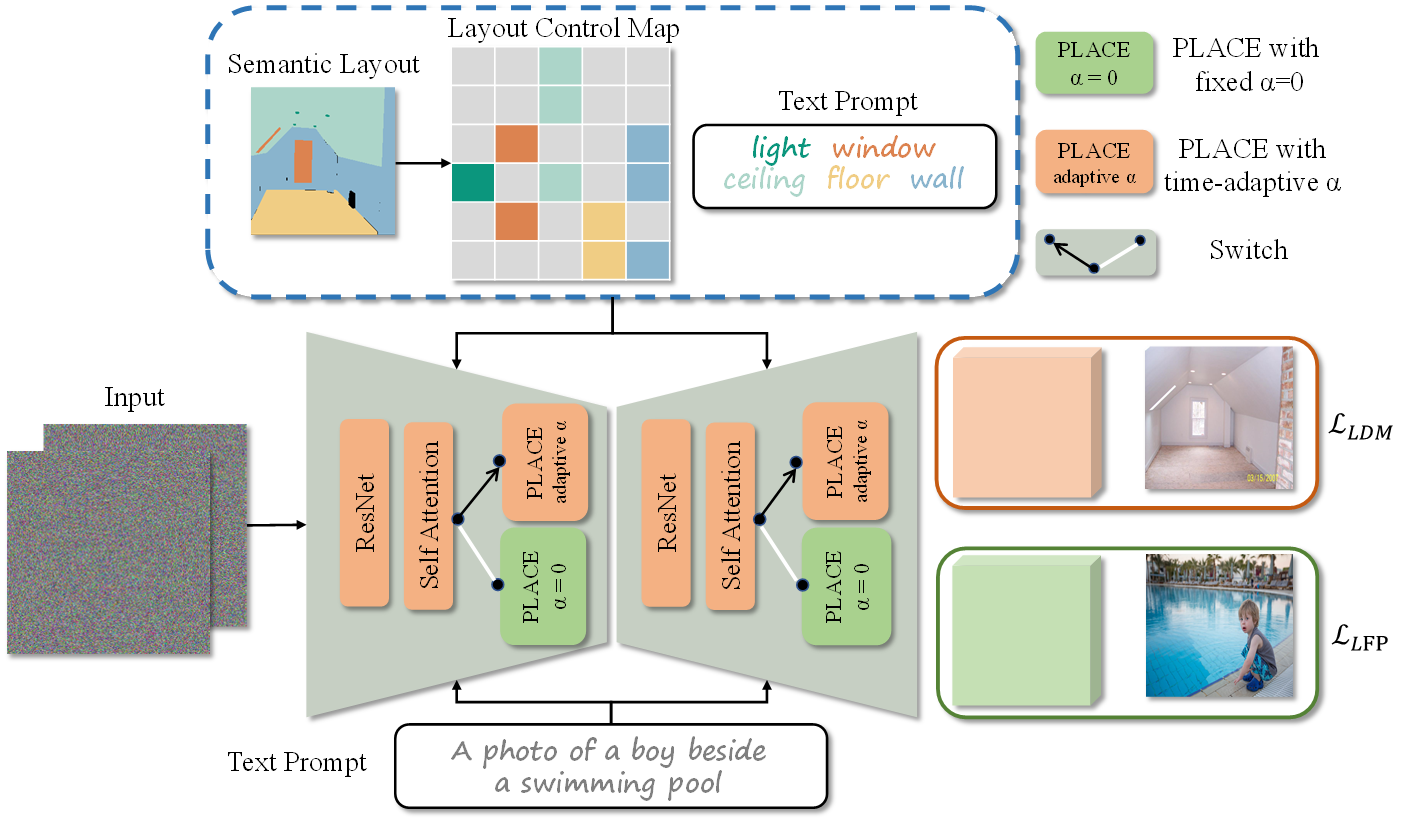}
\vspace{-10pt}
\caption{Fine-tuning with Layout-Free Prior Preservation Loss.
}
\label{fig:loss}
\vspace{-10pt}
\end{figure}

During each fine-tuning iteration, we sample a set of image-semantic layout pairs $<z_t, S, y, t>$ and a set of image-text pairs $<z_t', y', t'>$ from the Layout Free dataset. As depicted in Fig.~\ref{fig:loss}, for training data annotated with semantic layouts, we employ the PLACE with timestep-adaptive $\alpha$ to synthesize images and compute the semantic image synthesis loss $\mathcal{L}_{LDM}$. For image-text training data, we adopt the PLACE with fixed $\alpha(\alpha=0)$ to synthesize images and calculate the layout-free prior preservation loss $\mathcal{L}_{LFP}$. 

\section{More Ablation Study Results}
\label{sec:more_ablation}

\subsection{Layout Control Map}
\begin{figure}
\vspace{-4pt}
\centering
\includegraphics[width=.85\linewidth]{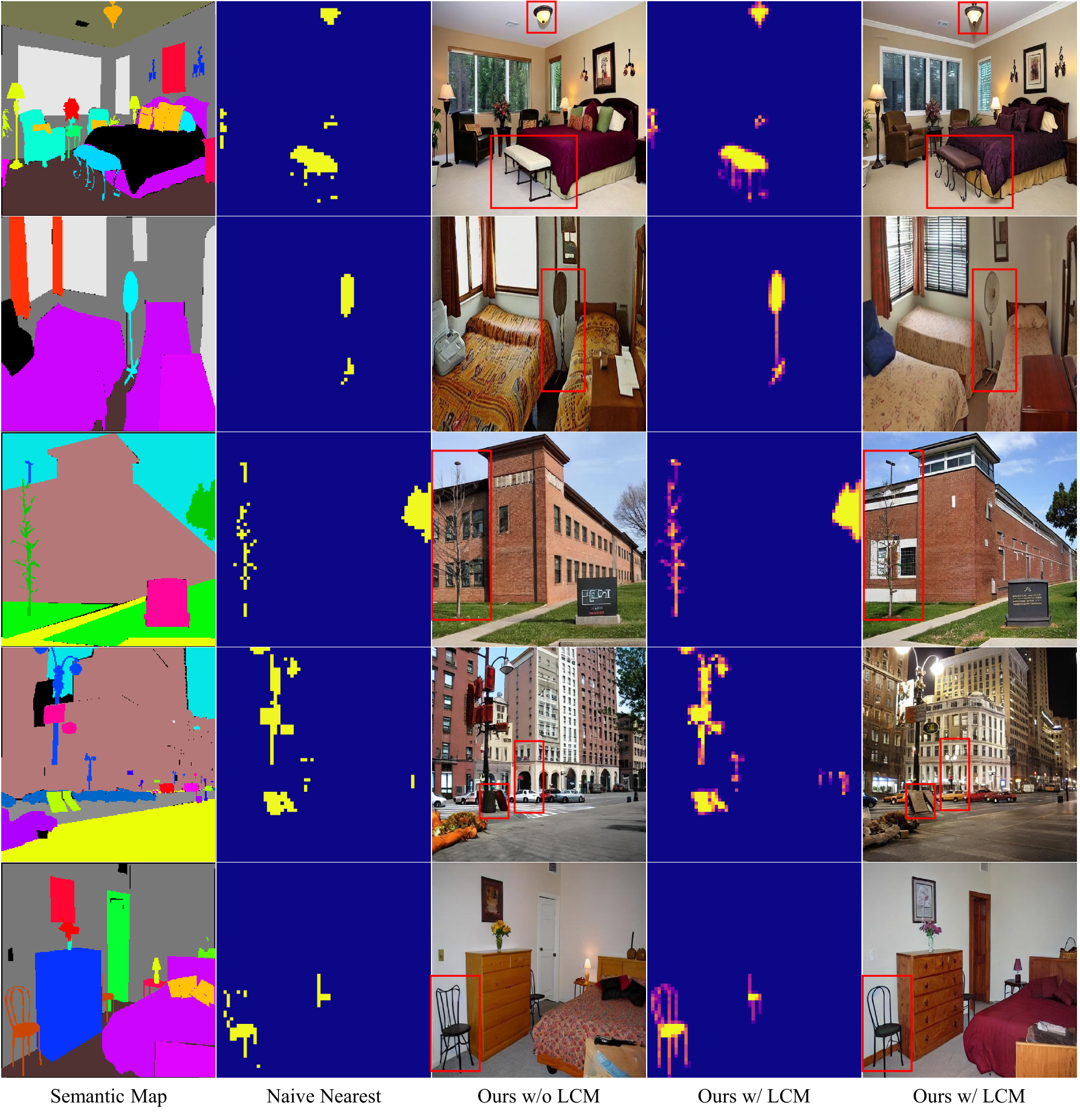}
\vspace{-10pt}
\caption{More Qualitative Ablation Comparisons on LCM. The $2nd$ and $4nd$ depict the layout representation for part of semantics.
}
\label{fig:lcm}
\vspace{-5pt}
\end{figure}

Fig.~\ref{fig:lcm} illustrates the impact of the Layout Control Map (LCM) on the generated images. The $2nd$ and $3rd$ columns, as well as the $4th$ and $5th$ columns, represent layout presentations ($64\times64$) without LCM and with the utilization of LCM, respectively, along with their corresponding synthesized images. It can be observed that LCM preserves more details in the low-resolution feature space, thus promoting faithful details and improved layout consistency.

\subsection{Adaptive $\alpha$ for fusion}
\begin{figure}
\vspace{-4pt}
\centering
\includegraphics[width=.85\linewidth]{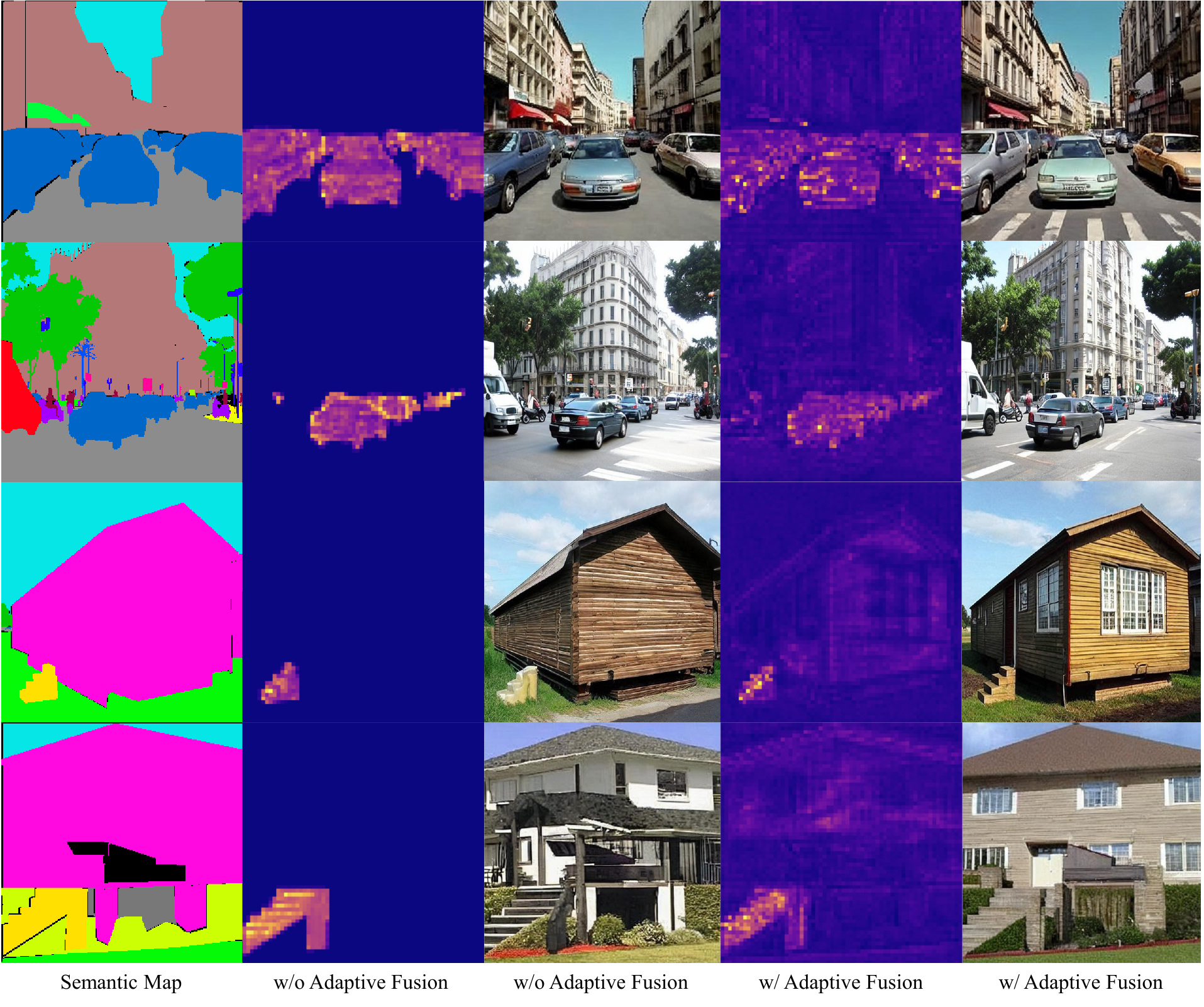}
\vspace{-10pt}
\caption{More Qualitative Comparisons on Adaptive Fusion. The $2nd$ and $4nd$ show the fusion map for part of semantics.
}
\label{fig:ada}
\vspace{-15pt}
\end{figure}

Fig.~\ref{fig:ada} displays the synthesis results with and without Layout-Semantic Adaptive Fusion. The $2nd$ and $4th$ columns represent the fusion maps corresponding to fixed alpha and adaptive alpha, respectively. It can be seen that adaptive fusion preserves the interactions of specified semantics (such as 'stairs' or 'car') on relevant semantic regions (such as 'house' or 'road'), resulting in the synthesis of more realistic details and higher visual quality.

\subsection{Semantic Alignment loss}
\begin{figure}
\vspace{-4pt}
\centering
\includegraphics[width=.85\linewidth]{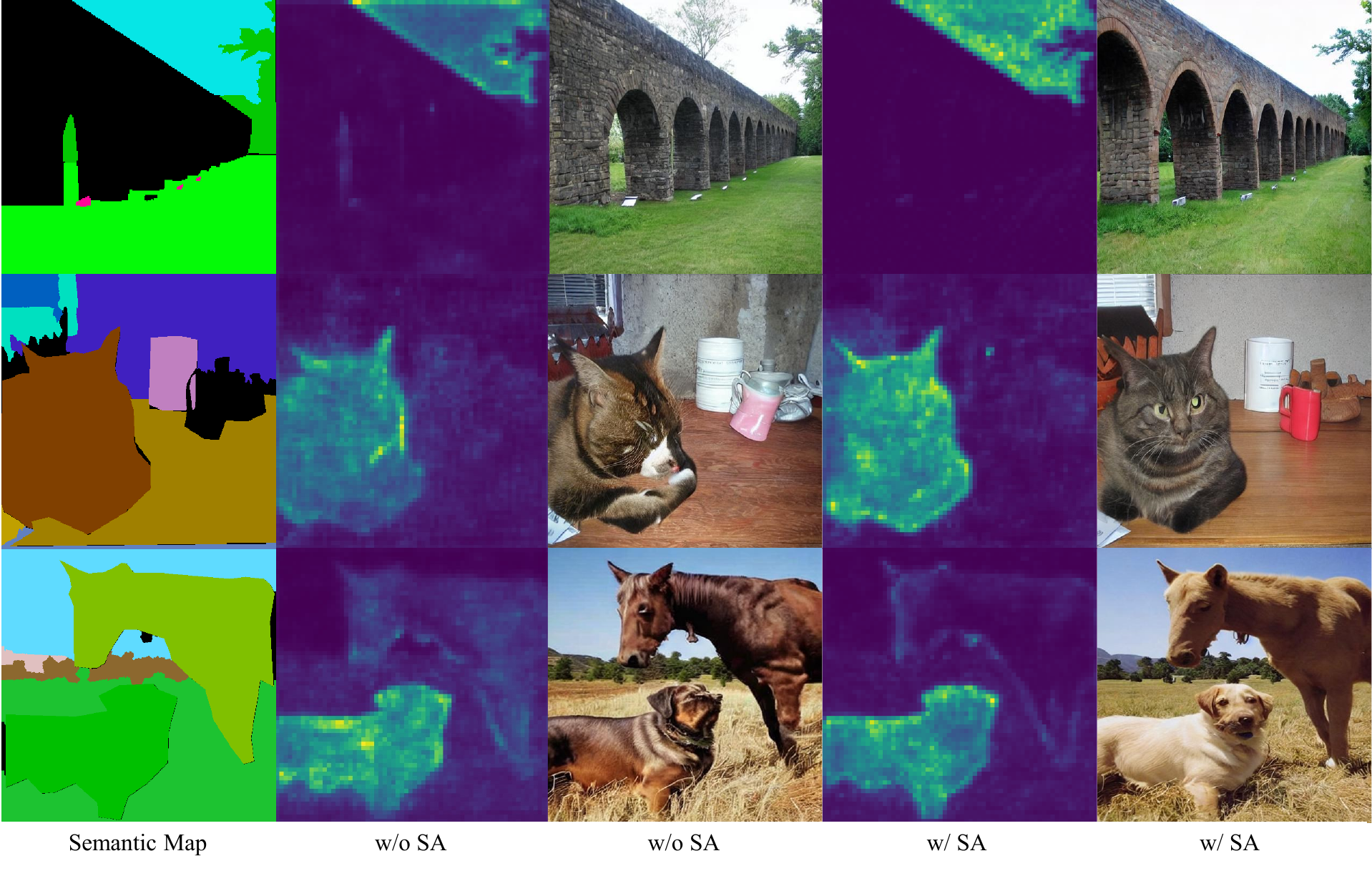}
\vspace{-10pt}
\caption{More Qualitative Comparisons on SA Loss. The $2nd$ and $4nd$ show the self-attention map for part of semantics.
}
\label{fig:sa}
\vspace{-18pt}
\end{figure}

Fig.~\ref{fig:sa} presents the results of the model fine-tuned with and without Semantic Alignment loss. The $2nd$ and $4th$ columns respectively show the self-attention maps of models fine-tuned without and with SA loss. It can be observed that the SA loss facilitates the interaction of image tokens within the same or related semantic regions (\eg 'sky', 'cat', and 'dog'), thereby improving the layout consistency and visual quality of synthesized images. 

\subsection{Layout Free Prior Preservation loss}
\begin{figure}
\vspace{-4pt}
\centering
\includegraphics[width=0.85\linewidth]{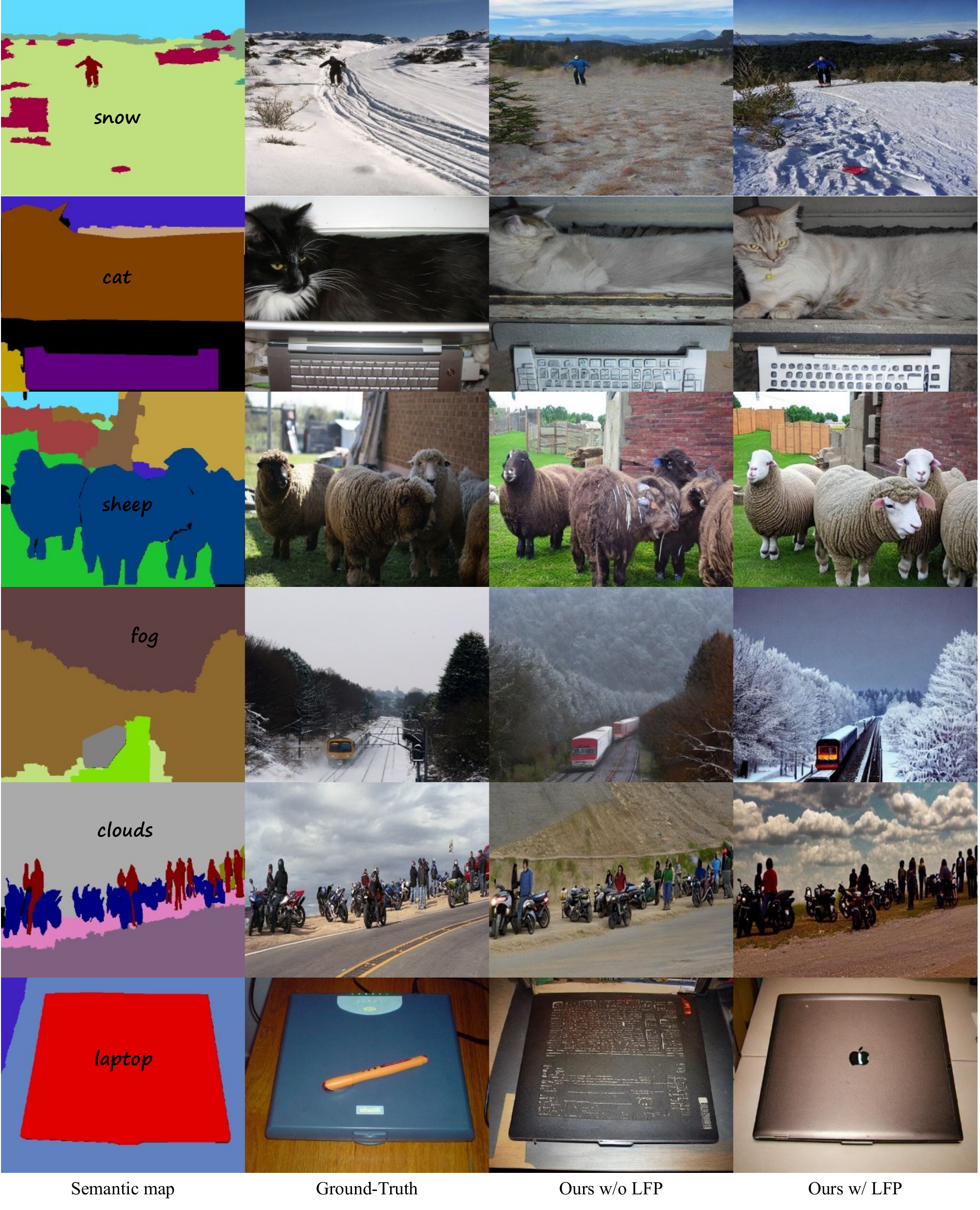}
\vspace{-10pt}
\caption{More Qualitative Ablation Comparisons on LFP Loss.
}
\label{fig:lfp}
\vspace{-10pt}
\end{figure}

\begin{figure*}[t]
\vspace{-1pt}
\centering
\includegraphics[width=.71\linewidth]{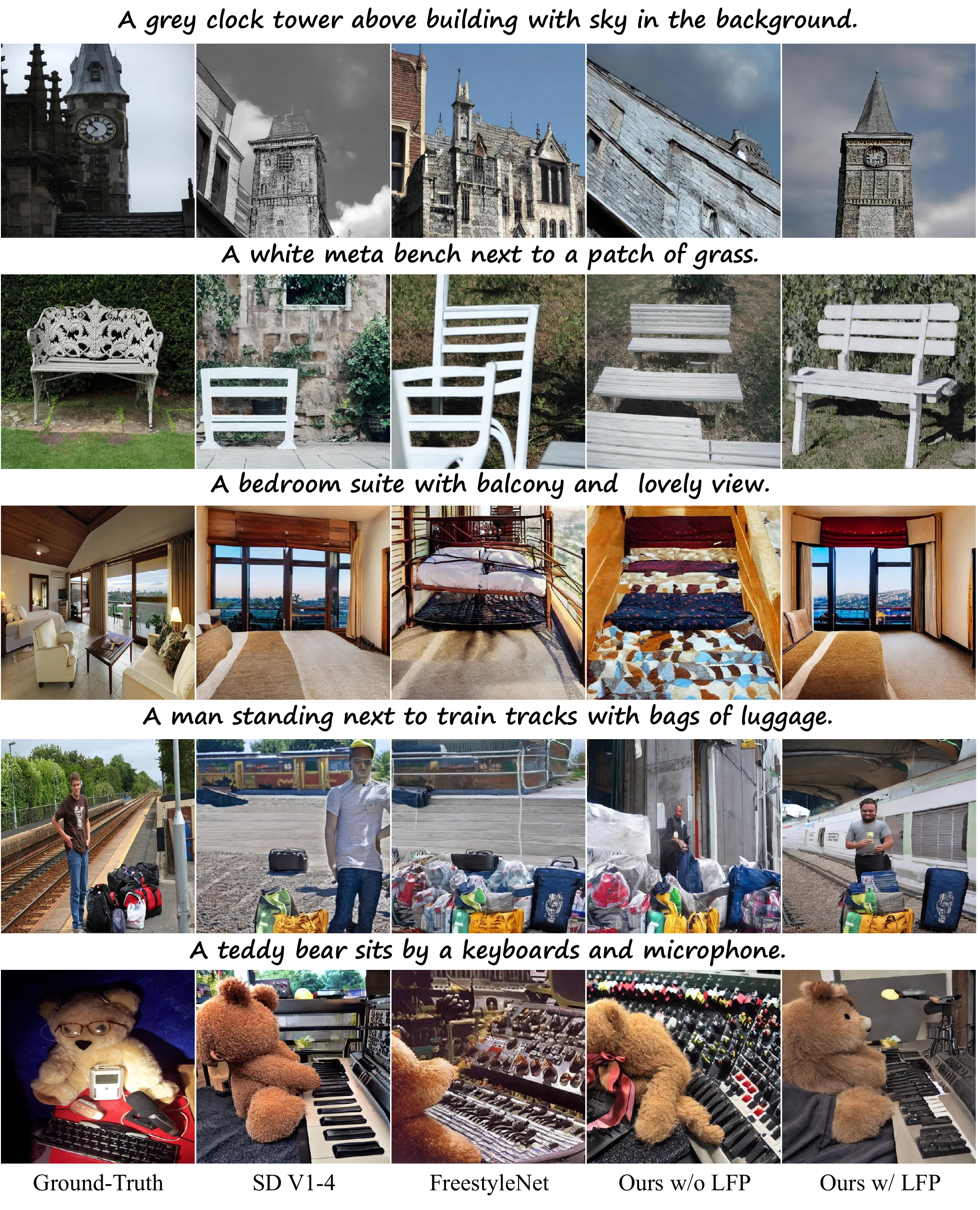}
\vspace{-10pt}
\caption{Comparison of Original Text-to-Image Synthesis results across different models.}
\label{fig:t2i}
\vspace{-15pt}
\end{figure*}

In this section, we first present more ablation qualitative comparison results (new object synthesis) on the Layout Free Prior Preservation loss in Fig.~\ref{fig:lfp}. Additionally, we validate the effectiveness of our layout-free prior preservation loss by assessing the original text-to-image synthesis capability of different models, as shown in Table~\ref{tab:t2i} and Fig.~\ref{fig:t2i}. 

Firstly, as observed from Fig.~\ref{fig:lfp}, the utilization of LFP Loss results in enhanced visual quality in the synthesis of semantic images. Notably, 'cat' in the $2nd$ row, 'sheep' in the $3rd$ row, and 'laptop' in the $6th$ row, all demonstrate improved visual results. Additionally, the semantic consistency of the results has been elevated, as shown in the $4th$ row with 'fog' and the $5th$ row with 'clouds'. These results collectively substantiate the effectiveness of the LFP Loss.

Then we assessed the original text-to-image synthesis capabilities of four models: Original Stable Diffusion V1-4~(SD V1-4), FreestyleNet, our model without using LFP~(Ours w/o LFP), and our model with LFP~(Ours w/ LFP). In this case, both our models and FreestyleNet were fine-tuned on the ADE20K dataset using SD V1-4 as the initial parameters.
We extract 1500 captions as input text prompts from the validation set of COCO-Stuff. During sampling, we employ 50 PLMS sampling steps with a classifier-free guidance scale of 2.
We calculate the FID and Text-Alignment scores between the synthesized images and ground truth, as shown in Table~\ref{tab:t2i}.  
It can be observed that the adoption of the LFP Loss significantly preserves the original text-to-image synthesis capability of the fine-tuned model. The FID decreases from 46.2 to 36.8, and Text-Alignment increases from 0.27 to 0.30, approaching the performance of the original model. This indicates the important role of LFP loss in preserving semantic concepts in the original model.
Furthermore, the qualitative comparisons in Fig.~\ref{fig:t2i} also indicate our LFP Loss helps preserve priors in the original model, resulting in the synthesis of images with improved semantic consistency and visual quality.

\begin{table}[!ht]
    \centering
    \begin{tabular}{c c c}
    \toprule[1.25pt]
        \textbf{Method} & \textbf{FID$\downarrow$}  & \textbf{Text-Alignment $\uparrow$} \\ \hline
        SD V1-4 & \textbf{34.4} & \textbf{0.31} \\ 
        FreestyleNet & 47.4 & 0.27 \\ 
        Ours w/o LFP & 46.2 & 0.27 \\ 
        Ours w/ LFP & \underline{36.8}  & \underline{0.30} \\ 
    \bottomrule[1.25pt]
    \end{tabular}
    \vspace{-5pt}
    \caption{Quantitative comparison of Text-to-Image Synthesis.}
    \label{tab:t2i}
    \vspace{-15pt}
\end{table}

\vspace{-1em}
\section{More Qualitative Results}
\label{sec:more_results}
\vspace{-0.5em}
\subsection{Additional Details of OOD Evaluation}
\vspace{-0.5em}
For quantitative evaluation of Out-Of-Distribution(OOD) synthesis, we conduct experiments from three aspects: new object, new attribute, and new style. We assess the generalization capability of the model fine-tuned on ADE20K.
For new object synthesis, we utilize the model to synthesize 5000 images in the validation set of COCO-Stuff. We compute the FID score of synthesized images and the mIoU of semantic classes exclusive to COCO-Stuff.
For new attribute synthesis, a total of 260 images are synthesized for six attributes: "\textit{brick} wall", "sky \textit{with rainbow}", "\textit{autumn} flora/tree/grass", "\textit{wooden} floor", "\textit{snowy} road", and "\textit{colorful} carpet". The text alignment between images and text prompts is computed with CLIP.
For new style synthesis, a total of 260 images are synthesized across eight different styles: "drawn by Van Gogh", "in oil painting", "in Minecraft", "full of graffiti", "in sketch", "in Monet style", "in anime" and "drawn by Picasso." The text alignment between images and text prompts is assessed by CLIP. Additionally, the semantic layouts used for evaluating new attribute and new style are both sampled from the ADE20K.

\subsection{Out-of-distribution Synthesis.} 
\begin{figure*}[t]
\vspace{-1pt}
\centering
\includegraphics[width=0.88\linewidth]{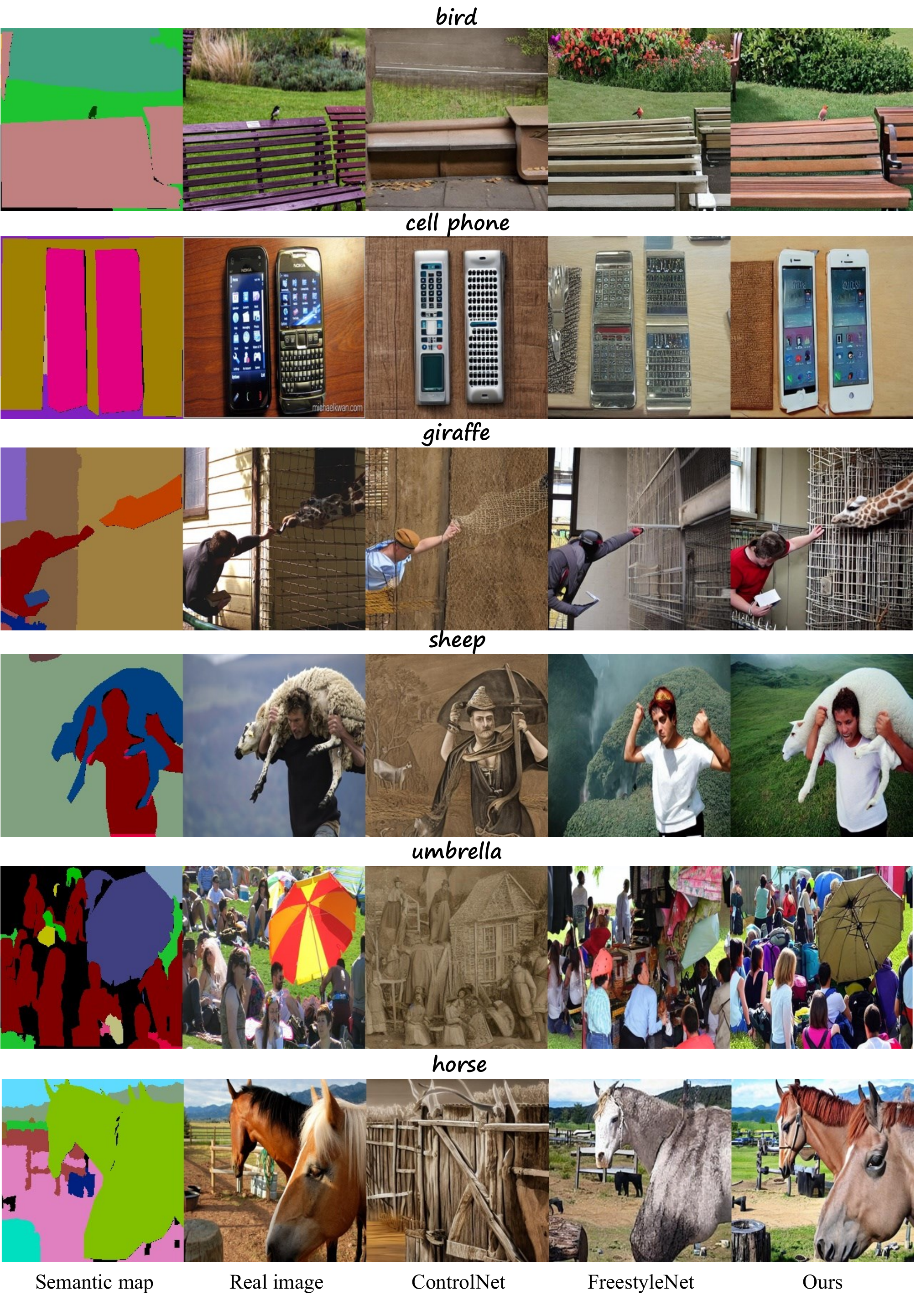}
\vspace{-10pt}
\caption{Visual comparisons on new object synthesis.}
\label{fig:obj}
\vspace{-15pt}
\end{figure*}

\begin{figure*}[t]
\vspace{-1pt}
\centering
\includegraphics[width=0.88\linewidth]{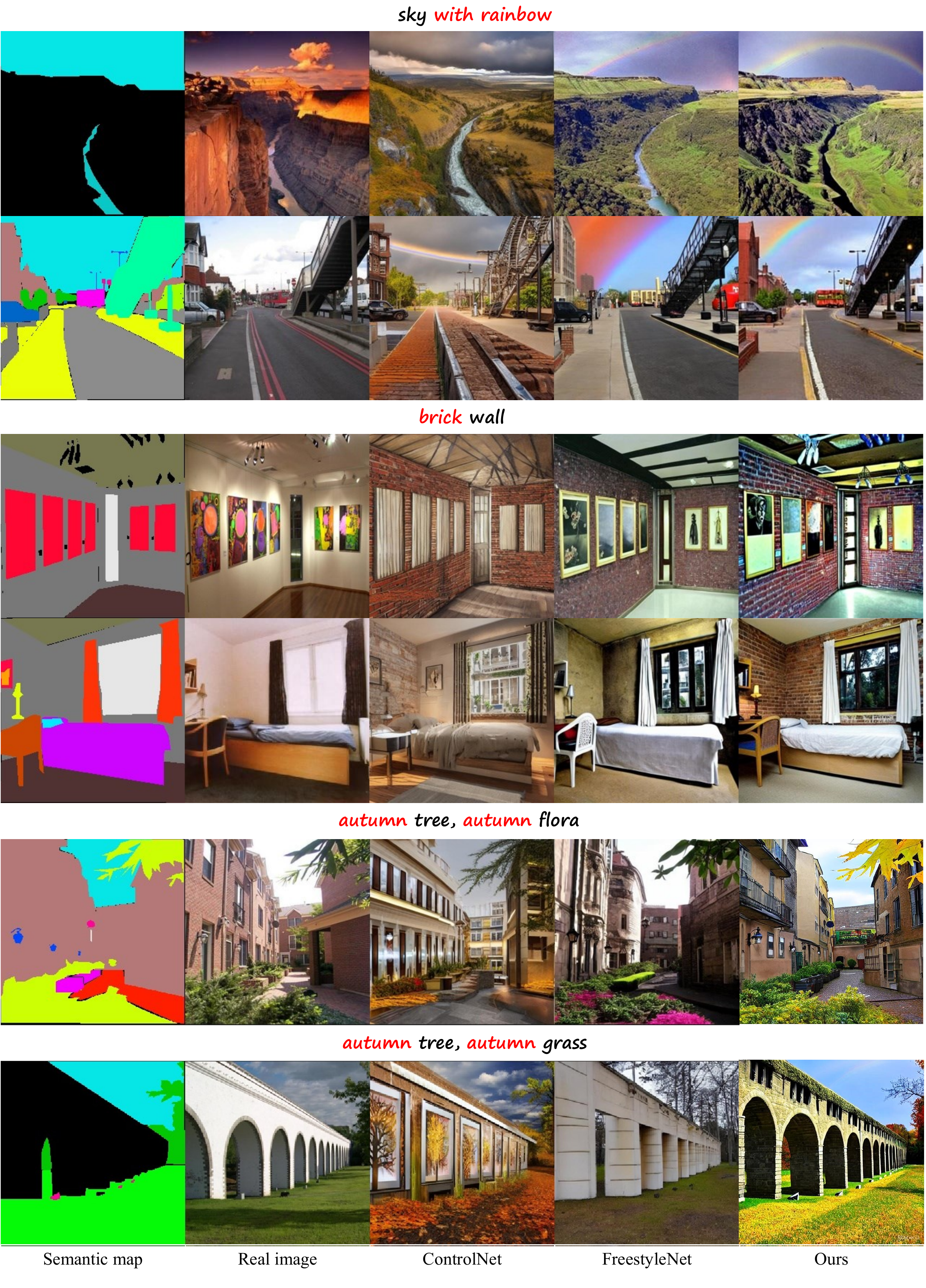}
\vspace{-10pt}
\caption{Visual comparisons on new attribute synthesis.}
\label{fig:attr}
\vspace{-15pt}
\end{figure*}

\begin{figure*}[t]
\vspace{-1pt}
\centering
\includegraphics[width=0.88\linewidth]{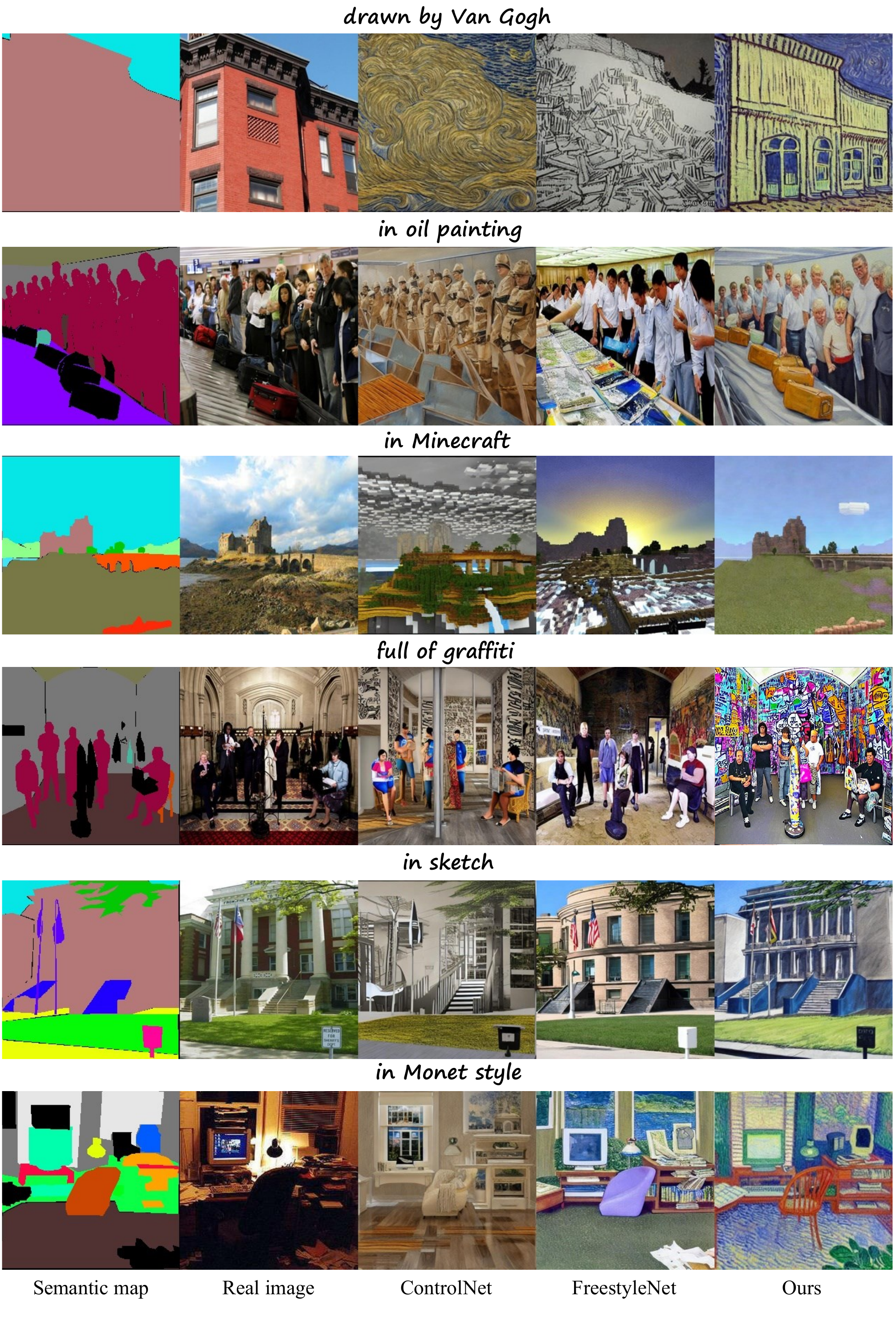}
\vspace{-10pt}
\caption{Visual comparisons on new style synthesis.}
\label{fig:sty}
\vspace{-15pt}
\end{figure*}

\vspace{-0.5em}
We show more visual comparisons with the competing methods on out-of-distribution synthesis. Fig.~\ref{fig:obj}, Fig.~\ref{fig:attr} and Fig.~\ref{fig:sty} respectively present the qualitative comparisons of the new object, new attribute, and new style, which are synthesized by a model fine-tuned on the ADE20K dataset. It can be observed that our approach synthesizes results that are more consistent with the provided text input.

\vspace{-0.5em}
\subsection{In-distribution Synthesis.}
\begin{figure*}[t]
\vspace{-1pt}
\centering
\includegraphics[width=1.\linewidth]{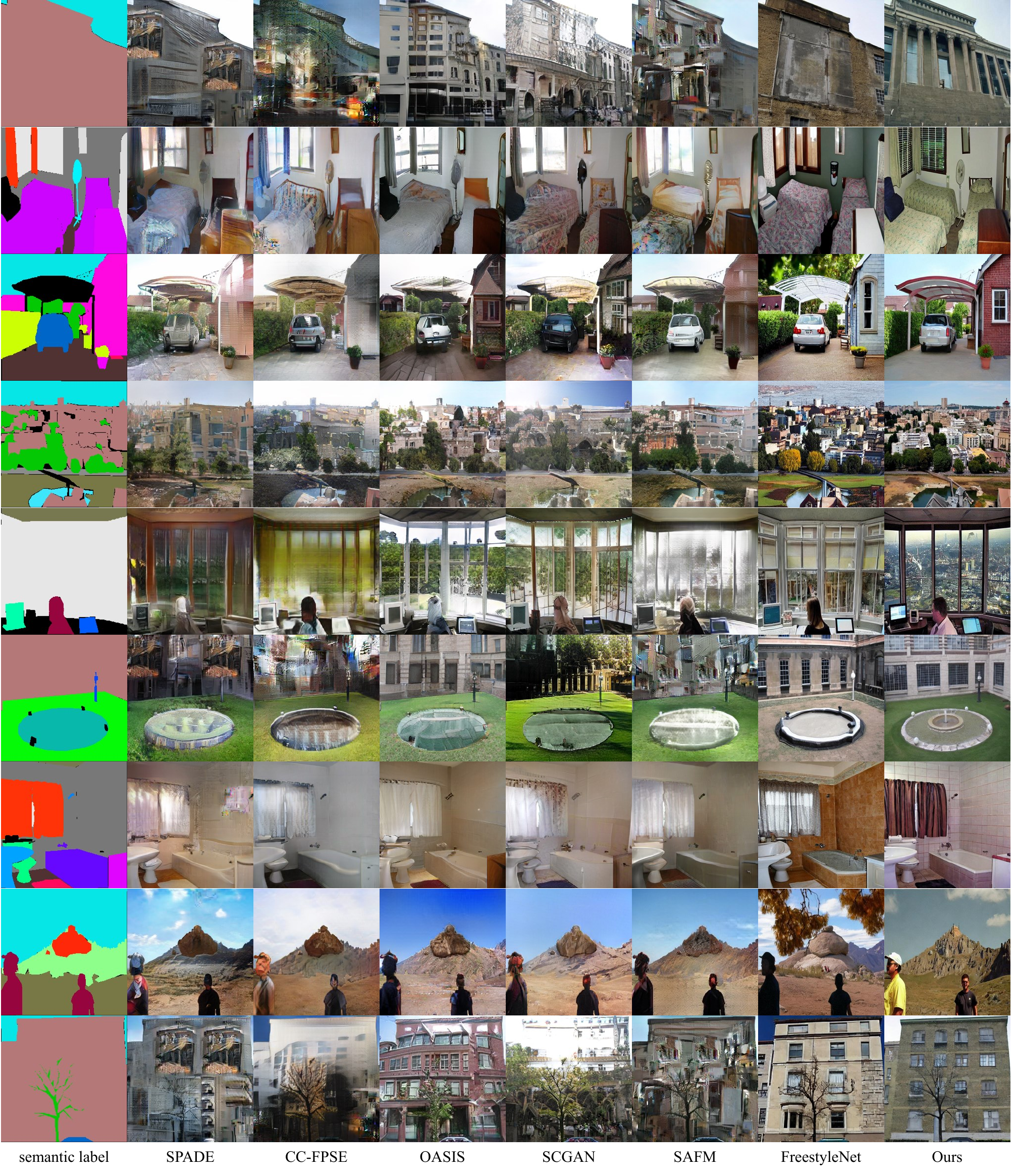}
\vspace{-10pt}
\caption{Visual comparisons on ADE20K dataset.}
\label{fig:ade}
\vspace{-15pt}
\end{figure*}

\begin{figure*}[t]
\vspace{-1pt}
\centering
\includegraphics[width=1.\linewidth]{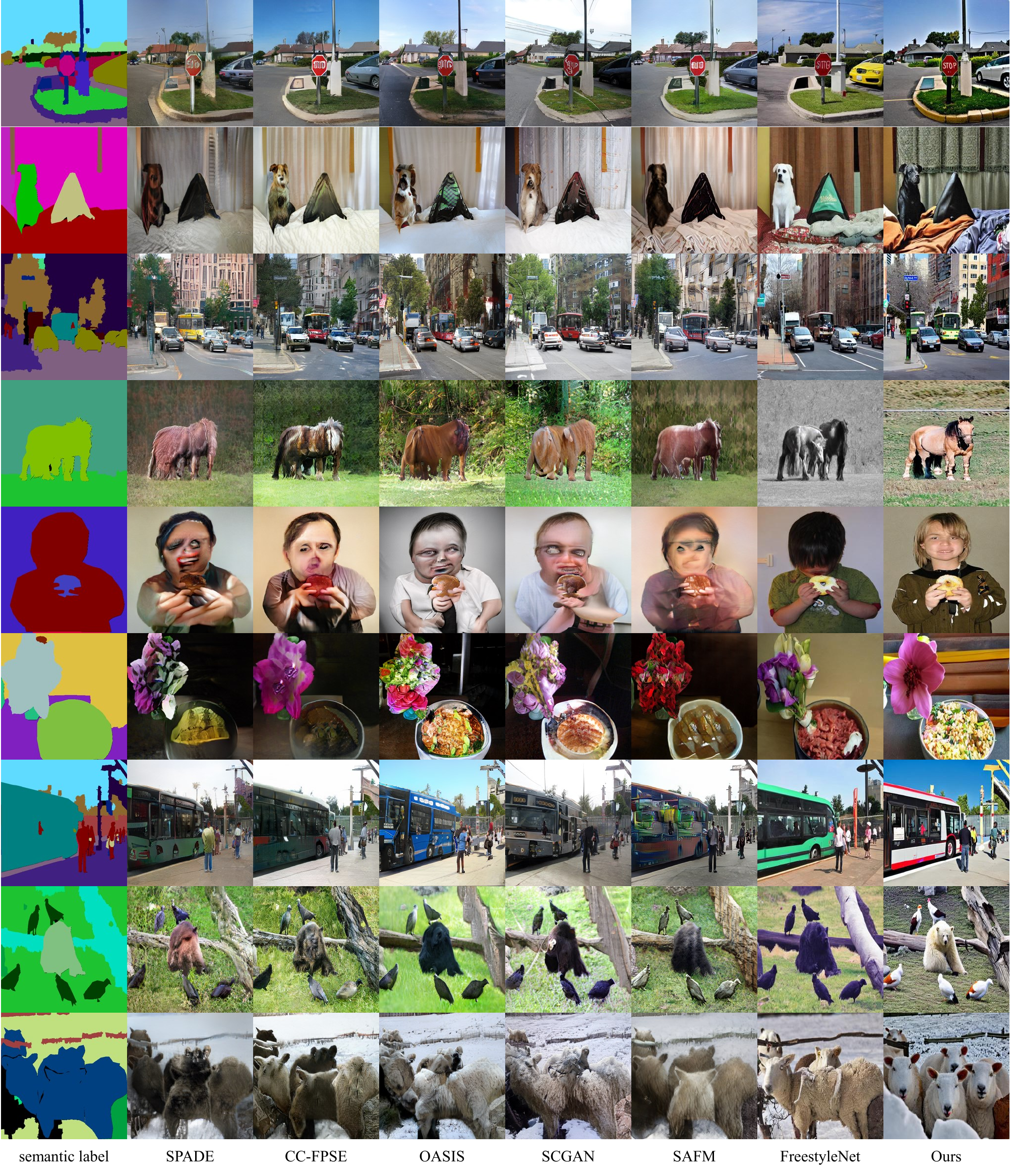}
\vspace{-10pt}
\caption{Visual comparisons on COCO-Stuff dataset.}
\label{fig:coco}
\vspace{-15pt}
\end{figure*}

\vspace{-0.5em}
Fig.~\ref{fig:ade} and Fig.~\ref{fig:coco} respectively illustrate more in-distribution qualitative comparisons on the ADE20K and COCO-Stuff dataset. It can be seen that our synthesized images are not only of high fidelity but also exhibit a strong alignment with the provided semantic layout in terms of finer details.

\vspace{-0.5em}
\section{Limitation}
\vspace{-0.8em}
\label{sec:limitation}

Although PLACE has made advancements in visual quality, semantic consistency, and layout alignment, there are still some limitations. Firstly, the inference speed of diffusion-based methods is still slower compared to that of GAN-based methods. On a V100 GPU, ControlNet, FreestyleNet, and PLACE require an average of approximately 7.5 seconds (s), 5.9 s, and 6.1 s, respectively, to synthesize an image using PLMS sampling for 50 steps. We believe that with the development of superior samplers and latent consistency models, this issue will be largely alleviated. Additionally, constrained by the capabilities of the pre-trained stable diffusion, when prompts for a single class are too long or contain uncommon tokens, the synthesized image may be inconsistent with the given text. Higher-performance text-to-image models may potentially ameliorate this issue.

% {
%     \small
%     \bibliographystyle{ieeenat_fullname}
%     \bibliography{main}
% }

\end{document}